%% file: main.tex
\documentclass[10pt,twocolumn,letterpaper]{article}

\usepackage[pagenumbers]{cvpr}

\newcommand{\lightr}{Light3R-SfM\xspace}
\newcommand{\mastr}{MASt3R\xspace}
\newcommand{\dustr}{DUSt3R\xspace}

\newcommand{\spannr}{Spann3R\xspace}
\newcommand{\ours}{SAIL-Recon\xspace}

\input{preamble}

\definecolor{cvprblue}{rgb}{0.21,0.49,0.74}
\usepackage[pagebackref,breaklinks,colorlinks,citecolor=cvprblue]{hyperref}
\usepackage[ruled,vlined,linesnumbered]{algorithm2e}
\usepackage{amsmath,amsfonts}
\title{SAIL-Recon: Large SfM by Augmenting Scene Regression with Localization
}

\author{%
Junyuan Deng\textsuperscript{1,2}\thanks{Equal contribution.}\quad
Heng Li\textsuperscript{1}\footnotemark[1]\quad
Tao Xie\textsuperscript{2,3}\quad
Weiqiang Ren\textsuperscript{2}\quad
Qian Zhang\textsuperscript{2}\\
Ping Tan\textsuperscript{1}\thanks{Corresponding authors.\hangindent=1.8em\hangafter=1\\E-mail: {pingtan@ust.hk}, {xiaoyang.guo@horizon.auto}}\quad
Xiaoyang Guo\textsuperscript{2}\footnotemark[2]
\\
\vspace{-0.4cm}
\and
\textsuperscript{1}{The Hong Kong University of Science and Technology}\quad
\textsuperscript{2}{Horizon Robotics}\quad
\textsuperscript{3}{Zhejiang University}\\
\vspace{3pt}\\
}
\begin{document}
\maketitle
\input{sec/0_abstract}    
\input{sec/1_intro}

\input{sec/2_related}

\input{sec/3_method}

\input{sec/4_exp}

\input{sec/5_conclusion}

{
    \small
    \bibliographystyle{ieeenat_fullname}
    \bibliography{main}
}
\input{sec/X_suppl}

\end{document}

%% file: preamble.tex
\usepackage[dvipsnames, table,xcdraw]{xcolor}
\usepackage{array}
\usepackage{multirow, makecell}
\usepackage{hhline}
\usepackage[most]{tcolorbox} 
\usepackage[export]{adjustbox}
\usepackage{siunitx}
\usepackage{rotating}
\usepackage{graphicx}
\usepackage{soul}
\definecolor{TableWorsttogood0}{RGB}{224,164,164}
\definecolor{TableWorsttogood1}{RGB}{231,178,172}
\definecolor{TableWorsttogood2}{RGB}{238,192,181}
\definecolor{TableWorsttogood3}{RGB}{182,215,168}
\definecolor{TableWorsttogood4}{RGB}{246,206,190}
\definecolor{TableWorsttogood5}{RGB}{252,220,198}
\definecolor{TableWorsttogood6}{RGB}{255,228,202}
\definecolor{TableWorsttogood7}{RGB}{254,232,204}
\definecolor{TableWorsttogood8}{RGB}{254,236,205}
\definecolor{TableWorsttogood9}{RGB}{253,240,207}
\definecolor{TableWorsttogood10}{RGB}{246,240,203}
\definecolor{TableWorsttogood11}{RGB}{227,230,189}
\definecolor{TableWorsttogood12}{RGB}{207,220,174}
\definecolor{TableWorsttogood13}{RGB}{187,210,160}
\definecolor{TableWorsttogood14}{RGB}{168,201,146}

\definecolor{TableDarkGreen}{RGB}{182,215,168}
\definecolor{TableLightGreen}{RGB}{207,234,215}
\definecolor{TableYellow}{RGB}{255,242,204}
\definecolor{TableRed}{RGB}{244,204,204}

\newcommand{\firstc}{\cellcolor{TableWorsttogood14}}
\newcommand{\secondc}{\cellcolor{TableWorsttogood12}}
\newcommand{\thirdc}{\cellcolor{TableWorsttogood10}}
\newcommand{\Simthree}{\ensuremath{\mathrm{Sim}(3)}\xspace}
\newcommand{\mr}{MASt3R\xspace}

\newcommand{\SLfour}{\ensuremath{\mathrm{SL}(4)}\xspace}
\definecolor{myemerald}{rgb}{0.753, 0.898, 0.804}
\definecolor{mylightgreen}{rgb}{0.894, 0.933, 0.745}
\definecolor{myyellow}{rgb}{0.996, 0.972, 0.780}

\newcommand{\TCBTableBest}[1]{\tcbox[on line,boxsep=0pt,left=0pt,right=0pt,top=0pt,bottom=0pt,colframe=white,colback=TableWorsttogood14,highlight math style={enhanced},extrude by=0.75pt]{#1}}
\newcommand{\TCBTableSecond}[1]{\tcbox[on line,boxsep=0pt,left=0pt,right=0pt,top=0pt,bottom=0pt,colframe=white,colback=TableWorsttogood12,highlight math style={enhanced},extrude by=0.75pt]{#1}}
\newcommand{\TCBTableThird}[1]{\tcbox[on line,boxsep=0pt,left=0pt,right=0pt,top=0pt,bottom=0pt,colframe=white,colback=TableWorsttogood10,highlight math style={enhanced},extrude by=0.75pt]{#1}}


%% file: sec/0_abstract.tex
\begin{abstract}
Scene regression methods, such as VGGT~\cite{wang2025vggt}, solve the Structure-from-Motion (SfM) problem by directly regressing camera poses and 3D scene structures from input images. They demonstrate impressive performance in handling images under extreme viewpoint changes. However, these methods struggle to handle a large number of input images. To address this problem, we introduce \ours, a feed-forward Transformer for large scale SfM, by augmenting the scene regression network with visual localization capabilities. 
Specifically, our method first computes a neural scene representation from a subset of anchor images. The regression network is then fine-tuned to reconstruct all input images conditioned on this neural scene representation. 
Comprehensive experiments show that our method not only scales efficiently to large-scale scenes, but also achieves state-of-the-art results on both camera pose estimation and novel view synthesis benchmarks, including TUM-RGBD, CO3Dv2, and Tanks \& Temples. We will publish our model and code. Code and models are publicly available at: \url{https://hkust-sail.github.io/sail-recon/}.

\end{abstract}

%% file: sec/1_intro.tex
\section{Introduction}
\label{sec:intro}

\begin{figure*}[htbp]
    \centering
    \includegraphics[width=1.0\linewidth]{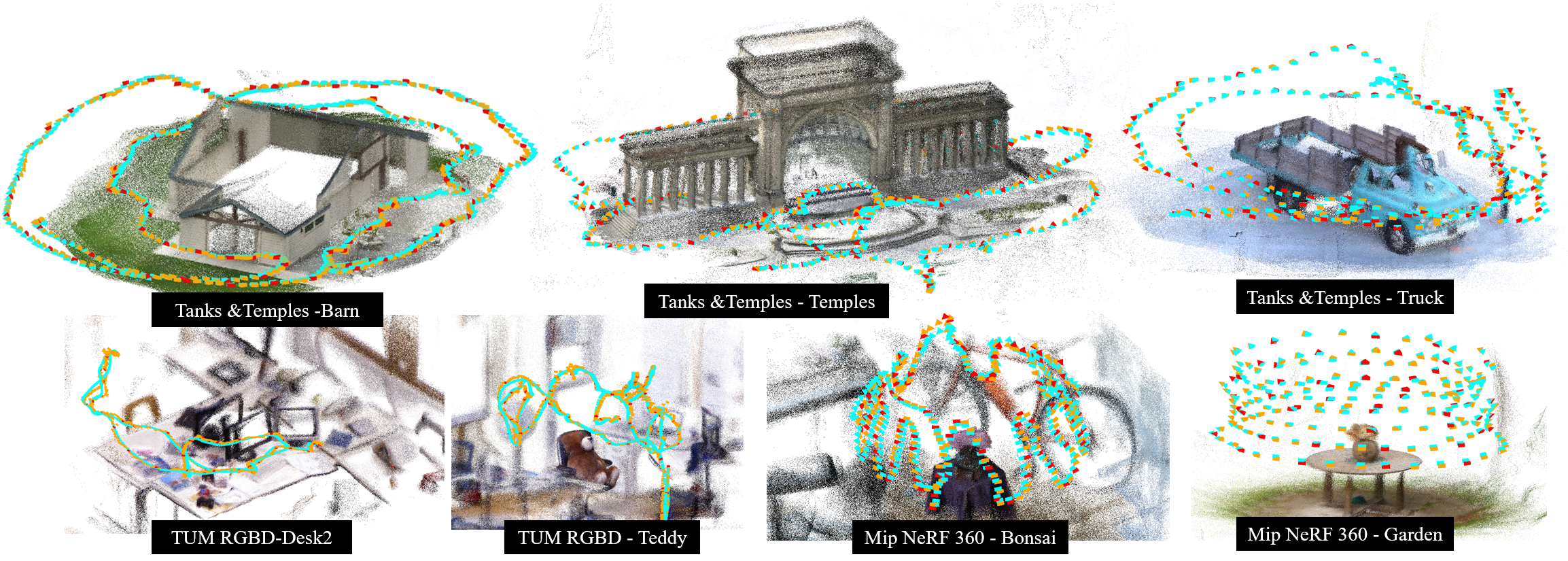}
    \caption{\textbf{Regressed Camera Poses and Point Clouds.} We visualize the camera poses and point clouds predicted by \ours across various datasets. COLMAP or ground-truth camera poses are shown as blue frustums, while regressed camera poses are shown in yellow, with red indicating anchor images. As illustrated, \ours estimates camera poses with accuracy comparable to COLMAP and produces high-quality, geometrically consistent point clouds.}
    \label{fig:visualization}
    \vspace{-3mm}
\end{figure*}

Structure-from-Motion (SfM) algorithms simultaneously estimate camera poses and scene structures from a collection of unordered images. This problem underlies many computer vision applications, such as novel view synthesis with NeRFs~\cite{mildenhall2021nerf, barron2021mip}, 3DGS~\cite{kerbl3Dgaussians}, multi-view stereo (MVS) reconstruction~\cite{yao2018mvsnet}, and visual localization~\cite{brachmann2017dsac}. Traditional SfM methods work either in incremental~\cite{ schoenberger2016sfm,snavely2006photo} or global~\cite{ cui2015global,pan2024glomap} approaches, which rely on crucial components such as feature detection~\cite{detone2018superpoint}, correspondence matching~\cite{sarlin2020superglue}, triangulation, and bundle adjustment~\cite{triggs1999bundle} for joint camera pose and scene structure optimization. However, these individual components are fragile to low-texture, blurred or repeated patterns, which could lead to catastrophic failures in the SfM process.

To overcome the limitation of conventional SfM methods, more recent works~\cite{wang2024dust3r,murai2025mast3r,yang2025fast3r,wang2025vggt} develop an end-to-end learning-based SfM pipeline to directly regress scene structures and camera poses from input images. \dustr~\cite{wang2024dust3r} pioneers this scene regression-based approach by training a Transformer~\cite{vaswani2017attention} to regress the scene coordinate maps (SCM) of two unposed images, which can be used to solve camera poses and correspondences.  
Some following works~\citep{yang2025fast3r,murai2025mast3r, elflein2025light3r} extend \dustr to multiple input images with 3D constraints, such as scene graph optimization and global alignment.
VGGT~\citep{wang2025vggt} develops the first large Transformer model to regress almost all 3D results end-to-end with a large dataset and multiple supervisions. Scene regression methods show impressive performance and robustness in handling unposed images with extreme viewpoint changes.

However, many scene regression methods, e.g., VGGT~\cite{wang2025vggt}, cannot scale up to videos or a large number of input images, as GPU memory usage increases quickly with more images.
Some methods~\cite {liu2025slam3r,wang2025continuous,wang2024spann3r} tackle video inputs using iteratively updated global memory tokens to fuse the features of each incoming frame, and regress scene coordinate maps conditioned on these memory tokens.
Others~\cite{elflein2025light3r,maggio2025vggt} divide the input video into segments, reconstruct each segment, and align different reconstructions using $Sim(3)$ or $SL(4)$. Both approaches suffer from pose drifting and are heavily dependent on subsequent global alignment to mitigate pose errors. 

On the other hand, existing scene regression methods ignore visual localization, a fundamental 3D vision task to solve the camera pose of a query image. Localization can facilitate scaling up an SfM system, a principle commonly employed in simultaneous localization and mapping (SLAM) systems, where mapping is only performed at keyframes and localization is applied to non-keyframes for better memory and computation efficiency. Our work seeks to augment scene regression with localization to scale it up in a similar spirit.

Most existing learning-based visual localization works~\cite{kendall2015posenet, brachmann2017dsac, brachmann2023accelerated, brachmann2024acezero} require time-consuming per-scene optimization and accurate camera pose annotations for the reference frames.
In contrast, we seek to fuse reconstruction and localization into a unified, annotation-free multitask framework for efficient scene regression. Specifically, given a large set of input images or video sequences, we first select a subset of anchor images to generate a global neural scene representation in a single forward pass, thereby avoiding the per-scene training required by previous localization methods. The neural scene representation serves as an implicit neural map of the scene, without relying on explicit 3D points or meshes. Subsequently, the neural scene representation, together with all remaining images, is fed into the same network to jointly recover scene coordinate maps and camera poses for each image. This approach allows for efficient reconstruction of thousands of images in just a few minutes. Unlike the localization process in SLAM systems, we regress map points for all images to produce a more complete 3D map that facilitates robust and dense reconstruction.

Our primary contributions are summarized as follows: 
\begin{itemize}
    \item We introduce \ours, a novel feedforward SfM method that generalizes neural scene regression to include localization, resulting in precise and robust reconstruction for thousands of input images in a few minutes. 
    \item We extract a neural scene representation from scene regression network, which serves as a global implicit map for localization.
    \item  We demonstrate through extensive experiments that \ours outperforms both traditional and learning-based baselines and achieves state-of-the-art results on SfM and visual localization benchmarks, including TUM-RGBD, CO3Dv2, and Tanks \& Temples.
\end{itemize}

%% file: sec/2_related.tex
\section{Related Works}
\label{sec:related_works}

\textbf{Geometric Structure-from-Motion (SfM)} is a classic computer vision problem~\cite{hartley2006multiple}, which aims to estimate camera poses and 3D scene structures from a collection of unposed images.
Traditional SfM solutions are categorized into two main approaches: Incremental SfM~\citep{snavely2006photo,agarwal2011building,frahm2010building,schoenberger2016sfm} initiates the reconstruction with a pair of images and progressively grows it by including images one by one; Global SfM methods~\cite{jiang2013global,wilson20141dsfm,cui2015global,pan2024glomap} determine the global pose of all images simultaneously by motion averaging. Both approaches rely on feature matching, triangulation, and bundle adjustment. Deep learning has significantly advanced these various components, especially in keypoint detection~\citep{yi_lift_2016, detone2018self, dusmanu2019d2, tyszkiewicz2020disk} and feature matching~\citep{sarlin2020superglue, chen2021learning, shi2022clustergnn, lindenberger2023lightglue}.
Beyond individual modules, several methods~\citep{tang2018ba, wei2020deepsfm,teed2018deepv2d, teed2021droid,brachmann2024acezero,smith2024flowmap,wang2024vggsfm} have explored end-to-end differentiable SfM by explicitly enforcing geometric constraints and minimizing reprojection or photometric errors.

\noindent
\textbf{Scene Regression-based SfM} recovers 3D structures and camera poses from uncalibrated images directly without explicitly enforcing geometric constraints. \dustr~\citep{wang2024dust3r} first employs a transformer model to predict the scene coordinate maps for a pair of images. Subsequent methods employ a global optimization step to expand its result to multiple images~\cite{wang2024dust3r,murai2025mast3r,wang2025continuous}. Recent advances have adapted \dustr to reconstruct multiple inputs directly~\cite{yang2025fast3r, tang2025mvdust3r,zhang2025flare,elflein2025light3r}, and to deal with video inputs with incremental reconstruction~\cite{wang2025continuous, liu2025slam3r, wang2024spann3r,maggio2025vggt,murai2025mast3r}. VGGT~\cite{wang2025vggt} takes this endeavor further, which addresses nearly all 3D vision tasks in a comprehensive end-to-end fashion with minimum inductive biases while utilizing extensive training data. However, these methods often face challenges when scaling to a large number of input images and may suffer from driftings, even when equipped with additional global alignment.

\begin{figure*}
    \centering
    \includegraphics[width=1.0\linewidth]{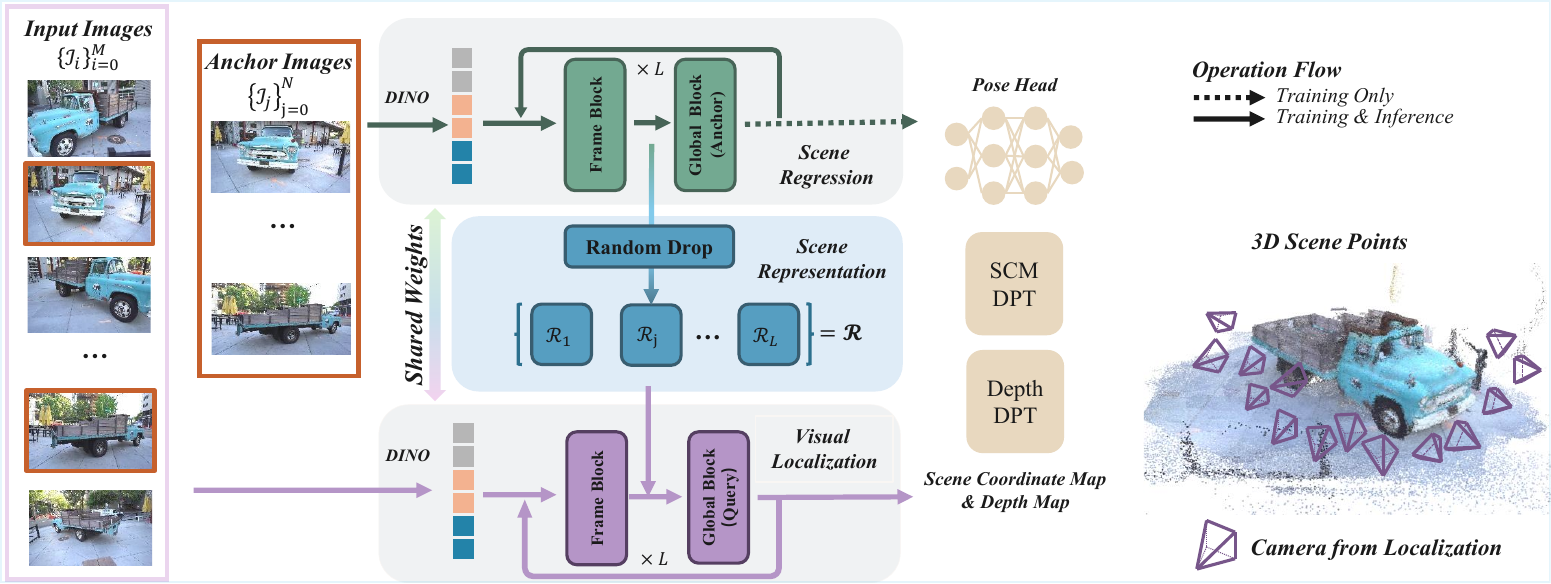}
    \caption{\textbf{Architecture Overview.} From a large set of unposed images, we first select a subset as anchor images, which are patchified by DINO~\cite{oquab2023dinov2} with appended camera tokens for scene regression.  The \textit{Scene Regression} block extracts a neural scene representation $\mathcal{R}$, which is then used by the \textit{Visual Localization} block to compute the camera poses and 3D scene points for all images. }
    \label{fig:pipeline}
    \vspace{-3mm}
\end{figure*}

\noindent
\textbf{Visual Localization} often relies on a 3D map with reference images of known camera poses.
Feature-based approaches extract 2D local features~\cite{detone2018superpoint,dusmanu2019d2,sun2021loftr,sarlin2021back} from a query image and match~\cite{sarlin2020superglue,lindenberger2023lightglue} them to 3D points and estimate the query image pose using a perspective-n-point (PnP) algorithm~\cite{gao2003complete}. For large-scale scenes, feature matching focuses on a subset of database images most relevant to the query, improving both accuracy and efficiency~\cite{humenberger2022investigating,rau2020predicting,sarlin2019coarse,sattler2017large}. 
Some learning-based approaches~\cite{kendall2015posenet,balntas2018relocnet,ding2019camnet,yang2022scenesqueezer,tang2023neumap,laskar2017camera,turkoglu2021visual} encode the map into a neural network that directly predicts the pose of the query image. 
The scene coordinate regression approaches~\cite{shotton2013scene,brachmann2017dsac,brachmann2021visual,brachmann2023accelerated} fit a scene-specific network to predict the 3D coordinates of the image pixels.
Most of the localization methods require expensive per-scene training of the localization network. Only a few methods~\cite{yang2019sanet,tang2021learning,turkoglu2021visual} estimate camera pose utilizing a network that simultaneously processes the 3D reconstruction and query image, thus bypassing the need for per-scene fitting.
Unlike previous methods, \ours circumvents expensive per-scene training by deriving a latent scene representation through scene regression, which is subsequently employed for visual localization.

%% file: sec/3_method.tex
\section{Method}

This section presents \ours, a unified framework for robust and efficient SfM with thousands of input images from various indoor or outdoor scenes, as shown in Fig.~\ref{fig:visualization}. To handle such scenes, we augment \textit{Neural Scene Regression} with \textit{Visual Localization}, which significantly reduces computational costs.
We first provide an overview of \ours in Sec.~\ref{sec:overview}. We then introduce the scene regression backbone in Sec.~\ref{sec:recon}, which estimates camera parameters and 3D structures from unordered input images. Building upon neural scene regression, Sec.~\ref{sec:subsample} details our approach for constructing a neural scene representation. This representation subsequently serves as the foundation for the visual localization introduced in Sec.~\ref{sec:reloc}. Sec.~\ref{sec:training} presents the training methodology, while Sec.~\ref{sec:refine} describes an optional refinement stage.

\subsection{Overview}
\label{sec:overview}
The pipeline of our approach is illustrated in Fig.~\ref{fig:pipeline}. The core contribution of \ours lies in the construction of a neural scene representation from a sparse subset of input images. Instead of relying on explicit geometric prior (e.g., point clouds, posed images), we leverage a neural representation that jointly encodes local visual features and global scene geometry from unposed images. This compact representation enables efficient visual localization across the entire image collection.

Motivated by the impressive performance of VGGT~\citep{wang2025vggt}, we adopt it as our backbone and augment its scene regression capability by the \textit{Visual Localization} block in Fig.~\ref{fig:pipeline}, to jointly compute a neural scene representation $\mathcal{R}$ as,
\begin{equation}
        (\mathcal{R},\{ T_i, K_i, D_i, S_i\}_{i=1}^M) = \mathcal{T}_{\theta}(\{\mathcal{I}_i\}_{i=1}^M),
        \label{equ:sceneregression}
\end{equation}
where $\{\mathcal{I}_i\}_{i=1}^M$ is the unordered input image set, $M$ is the total number of input images. $\mathcal{T}_{\theta}$ is a transformer-based network with self-attention blocks parameterized by $\theta$, and $\mathcal{R}$ is the neural scene representation for visual localization. The camera pose of $\mathcal{I}_i$ is specified by the extrinsic and intrinsic matrices $T_i\in \mathbb{R}^{4\times4}$ and $K_i\in \mathbb{R}^{3\times3}$. Furthermore, $D_i\in \mathbb{R}^{H\times W}$ and $S_i \in \mathbb{R}^{H\times W\times3}$ represent the depth map and the scene coordinate map (SCM) for $\mathcal{I}_i$, respectively.

The formulation in Eq.~\ref{equ:sceneregression} is memory and time consuming. It typically cannot handle more than 100 input images on consumer GPUs. 
To address this, we use a subset of images, called \textit{Anchor Images} in Fig.~\ref{fig:pipeline}, to compute the neural scene representation $\mathcal{R}$. Without losing generality, we uniformly sample $N\in[50,100]$ anchor frames $\{\mathcal{I}_i\}_{i=1}^{N}$ from the full set $\{\mathcal{I}_i\}_{i=1}^{M}$, where $M$ often exceeds 1,000 in large-scale scenes. 
As illustrated by the \textit{Visual Localization} block in Fig.~\ref{fig:pipeline}, once we have the scene representation $\mathcal{R}$, we augment the scene regression network $\mathcal{T}_\theta$ to enable localization of a query image $\mathcal{I}^q$ as,
\begin{equation}
    \{T^q,K^q,D^q,S^q\} = \mathcal{T}_{\theta}(\mathcal{I}^q, \mathcal{R}),
    \label{equ:efficient_reloc}
\end{equation}
where $T^q$ and $K^q$ are the extrinsics and intrinsics matrices, and $D^q$ and $S^q$ are the depth and scene coordinate maps of the query image $\mathcal{I}^q$. 
$\mathcal{T}_{\theta}$ is the network for both scene regression and localization. We will provide details in Sec~\ref{sec:reloc}.

Note that we estimate camera parameters and 3D maps of anchor frames during \textbf{TRAINING ONLY}, as indicated by the \textit{Operation Flow} in Fig.~\ref{fig:pipeline}. 
During inference, we first extract $\mathcal{R}$ from the anchor frames and then process all input images conditioned on $\mathcal{R}$. Although all input images could be reconstructed in a single forward pass in principle, we process them in batches due to GPU memory constraints.

\subsection{Neural Scene Regression}
\label{sec:recon}
The scene regression network takes a set of unposed anchor images $\{\mathcal{I}_i\}_{i=1}^N$ as input. Each anchor image $\mathcal{I}_i$ is fed into DINOv2~\citep{oquab2023dinov2} to extract patchified feature tokens $t^{\mathcal{I}_i}\in \mathbb{R}^{K\times C}$, where $K=(W/14)\times (H/14)$. We follow VGGT~\cite{wang2025vggt} to augment $t^{\mathcal{I}_i}$ with an additional camera token $t^{\mathcal{I}_i}_g\in\mathbb{R}^{1\times C}$ and four register tokens $t^{\mathcal{I}_i}_r\in\mathbb{R}^{4\times C}$.
The tokens from all anchor frames $\{t^{\mathcal{I}_i}\}$ are subsequently processed through $L=24$ layers of frame-wise and global self-attention as,
\begin{align}
     \{t_{j}^{'\mathcal{I}_{i}}\} &= \text{attn}_{j}^{\text{frame}}(\{t_{j}^{\mathcal{I}_{i}}\}),\\
     {t_{j+1}^{\mathcal{I}}} &= \text{attn}_{j}^{\text{global}}(t_{j}^{'\mathcal{I}}), \label{equ:global_attn}
\end{align}
where $j$ indexes the attention layers and $\{t_{j}^{'\mathcal{I}_{i}}\}$ is the output of frame-wise attention for $\mathcal{I}_i$. $t_j^{'\mathcal{I}}= [ \{t^{'\mathcal{I}_i}_j\}_{i=1}^{N}]$ is the concatenation of all image tokens in the $j$-th layer. The global attention result $t_{j+1}^{\mathcal{I}}$ is then split along the view dimension into $\{ t_{j+1}^{\mathcal{I}_{i}} \}$ for the frame-wise attention in the next layer.

The image tokens $\{ t^{\mathcal{I}_i}_{L} \}$ in the last layer are then fed into DPT heads~\cite{ranftl2021vision} to predict depth maps $D_i$ and scene coordinate maps $S_i$ for the input image $\mathcal{I}_i$ as follows,
\begin{equation}
    \{D_i,C^D_i,S_i,C^S_i\} = \text{DPT}(\{t^{\mathcal{I}_i}_{L}\}),
\end{equation}
where $C_i^D$ and $C_i^S$ are the confidence maps of depth and scene coordinate maps, respectively.

The camera tokens $t^{\mathcal{I}_i}_{g}$ associated with $\mathcal{I}_i$ on the last layer are processed by a camera head to estimate the intrinsic and extrinsic camera parameters as follows,
\begin{equation}
    \{T_i,K_i\} = \text{PoseHead}(\{t^{\mathcal{I}_i}_{g}\}).
\end{equation}

\subsection{Neural Scene Representation }
\label{sec:subsample}

Scene representation is essential for visual localization.
An effective scene representation should encode global 3D scene structure and facilitate correspondences between 3D map points and 2D image pixels.
A straightforward choice is to use tokens $t_{L}^{\mathcal{I}}$ in the final attention layer as scene representation, as these tokens can be used to recover camera poses and dense 3D points map.
In other words, we could set $\mathcal{R} = t_{L}^{\mathcal{I}}$ and then estimate the pose and 3D structure of a query image $\mathcal{I}^q$ by Eq.~\ref{equ:efficient_reloc}.
However, the significant discrepancy between 2D and 3D feature tokens makes it difficult for the network $\mathcal{T}_{\theta}$ to correlate these feature tokens, even $\mathcal{T}_{\theta}$ contains a large number of parameters. We empirically find that this design leads to suboptimal results.

Ideally, the scene representation should effectively bridge the gap between 2D and 3D features. Inspired by the fact that the network $\mathcal{T}_{\theta}$ takes a set of 2D images as input and progressively enhances them to compute 3D structures, we extract intermediate feature tokens from each attention layer of $\mathcal{T}_\theta$ to form the scene representation $\mathcal{R}$. In this way, our scene representation $\mathcal{R}$ captures the gradual evolution from 2D appearance features to 3D coordinate descriptors. Specifically, we compute our scene representation as,
\begin{equation}
\mathcal{R} = [\{\Theta(t_j^{'\mathcal{I}})\}_{j=1}^{L}],
\label{equ:scenerepresentation}
\end{equation}
where $\Theta$ denotes a downsampling operation applied to intermediate feature tokens $t_j^{'\mathcal{I}}$ to keep the scene representation compact. Specifically, for the anchor frame feature token $t_{j}^{'\mathcal{I}}\in \mathbb{R}^{N\times K\times C}$, we randomly select a ratio $r\in[0.2,1.0]$ to sample a subset of tokens in each anchor frame to control the total size of $\Theta(t^{'\mathcal{I}}_j)\in\mathbb{R}^{(N\times \lfloor r\times K \rfloor)\times C}$ in a reasonable range during training. At testing time, we could choose an appropriate $r$ based on the numer of the anchor frames to balance between accuracy and efficency.

\subsection{Neural Visual Localization}
\label{sec:reloc}

As described in Eq.~\ref{equ:global_attn}, the global attention block in $\mathcal{T}_{\theta}$ aggregates tokens among all anchor frames via self-attention.
For visual localization, given a query image $\mathcal{I}^q$, 
we introduce an attention mask in the global attention blocks, which essentially allows us to compute the cross attention between the query image $\mathcal{I}^{q}$ and the scene representation $\mathcal{R}_j=\Theta(t_j^{'\mathcal{I}})$:
\begin{equation}
    t^{q}_{j+1} 
    =\text{attn}^{\text{global}}_j( \{t^{q}_{j}, \mathcal{R}_j \} ).
    \label{equ:global_attn_query}
\end{equation}
Specifically, tokens from query frames cannot attend to tokens from other query frames; they can only attend to tokens within the same frame and to the scene representation.
More details are provided in the Supplementary. 

During inference of visual localization, one choice is to compute the camera parameters by applying the PnP algorithm~\cite{gao2003complete} with the scene coordinate map $S^q$ from the DPT head. However, it takes a long time for DPT to up-sample tokens to a high-resolution scene coordinate map. To accelerate pose estimation, we leverage the pose head, which takes only a few camera tokens $t_g^{\mathcal{I}_i}$ as input and directly regresses the camera pose. 

\begin{equation}
     \{T^q,K^q\} = \text{PoseHead}(t_g^q,\{t^{\mathcal{I}_i}_{g}\}) ,
\end{equation}
where $t_g^q$ and $\{ t^{\mathcal{I}_i}_{g} \}$ are the camera tokens of the query image and the anchor images, respectively. The attention mask is also applied in the pose head. Thus, we can formulate our visual localization for all images as follows,
\begin{equation}
    \{ T_i, K_i, D_i, S_i\}_{i=1}^M = \mathcal{T}_{\theta}(\{\mathcal{I}_{i}\}_{i=1}^{M},\mathcal{R}).
\end{equation}
Note that the anchor images are also processed by this localization block. The scene regression block only extracts the neural scene representation from anchor images.

\subsection{Training}
\label{sec:training}
\noindent
\textbf{Training Losses.} During training, we split the input images into the anchor image set and the query image set. We could forward all images in the anchor and query set in a single pass. To this end, we train the \ours model $\mathcal{T}_{\theta}$ end-to-end using a multitask loss on each frame similar to VGGT~\citep{wang2025vggt} as,
\begin{equation}\label{eq:training_loss}
\mathcal{L}
=
\mathcal{L}_\text{camera}
+ \mathcal{L}_\text{depth}
+ \mathcal{L}_\text{scm}.
\vspace{-2mm}
\end{equation}

The camera pose loss $\mathcal{L}_\text{camera}$ compares the predicted camera parameters $\hat{g}_i$ with the ground truth $g_i$ as 
\vspace{-2mm}
\begin{equation}
    \mathcal{L}_\text{camera} = \sum_{i=1}^N 
\left \| \hat{g}_i - g_i 
\right \|_1,
\label{equ:camera_loss}
\vspace{-2mm}
\end{equation}
where $g_i=[\mathbf{q},\mathbf{t},\mathbf{f}]$ includes the camera orientation encoded in a quaternion $\mathbf{q}$, the translation vector $\mathbf{t}$, and the field of view $\mathbf{f}$. We assume that the principal point is at the image center.
The depth loss $\mathcal{L}_\text{depth}$ follows \dustr~\citep{wang2024dust3r} to measure the discrepancy between the predicted depth $\hat{D}_i$ and the ground truth depth $D_i$ with a predicted uncertainty map $\hat{C}_i^D$.
We follow~\citep{wang2025vggt} to add a gradient-based smooth term on the depth loss, 
\vspace{-1mm}
{\small
\begin{equation}
        \mathcal{L}_\text{depth}
=
\sum_{i=1}^N
 \| C_i^D \odot (\hat{D}_i - D_i) \| + 
\| ({\small\nabla} \hat{D}_i - {\small\nabla} D_i) \|
- \alpha \log C_i^D,
\nonumber
\end{equation}
}
\vspace{-1mm}
where $\odot$ is an element-wise product. The loss of the scene coordinate map is defined similarly as,
{\small
\begin{equation}
\mathcal{L}_\text{scm}
=
\sum_{i=1}^N
\| C_i^S \odot (\hat{S}_i - S_i) \| + \| ({\small\nabla} \hat{S}_i - {\small\nabla}  S_i) \|
- \alpha \log C_i^S
.\nonumber
\end{equation}
}

\noindent
\textbf{Coordinate Normalization.}
During training, we randomly select an image $\mathcal{I}_{r}$ among the anchor images as the reference frame. 
We then compute the average Euclidean distance from the camera center of $\mathcal{I}_{r}$ to the 3D points of all anchor frames. 
We use this scale to normalize the 3D scene and the associated depth and scene coordinate maps.

\subsection{Post Refinement}
\label{sec:refine}
Employing a post-refinement step, such as bundle adjustment (BA)~\cite{triggs1999bundle}, can enhance reconstruction accuracy, particularly when the reconstruction is inferred directly from the network.  
Previous methods rely on global alignment using depth maps via gradient descent~\cite{wang2024dust3r,murai2025mast3r,wang2025continuous}, while others use bundle adjustment with pixel correspondences~\cite{wang2025vggt}.  
Both methods are time-consuming and scale poorly to a large number of input images. Since our method provides accurate camera poses in a global coordinate system, we adopt BARF-like methods~\cite{lin2021barf,nerfstudio} to optimize camera poses by minimizing a rendering loss. 
Although this optimization is weaker than bundle adjustment or global alignment, it scales to more than 10K frames and takes only $2$-$10$ minutes.
Note that this post-optimization is optional, since the camera poses and scene structures regressed by \ours yield strong results for many applications. Further details are provided in Sec.~\ref{sec:exp}.

%% file: sec/4_exp.tex
\input{tables/table_tnt_pose}

\section{Experiment}
\label{sec:exp}

\noindent
In this section, we present a comprehensive evaluation across diverse datasets, covering a wide range of scenarios. We also perform detailed ablation studies to identify the key factors determining the performance of our model. Additional implementation details for both training and inference are provided in the supplementary material.

\input{tables/table_tum}

\noindent
\textbf{Training Details.}
Our training is similar to the setup of VGGT~\citep{wang2025vggt}. We train our model by fine-tuning VGGT~\cite{wang2025vggt} pre-trained checkpoints. For each batch, we sample $4$–$48$ images, randomly designating $2$–$24$ as anchor frames and treating the remaining images as query frames. 
We train 30K iterations on 16 NVIDIA A800 GPUs, which takes about four days.
We use a diverse mixture of synthetic and real-world datasets, including Co3Dv2~\cite{reizenstein21co3d}, BlendMVS~\cite{yao2020blendedmvs}, DL3DV~\cite{ling2024dl3dv}, MegaDepth~\cite{li2018megadepth}, WildRGB~\cite{xia2024rgbd}, ScanNet++~\cite{yeshwanth2023scannet++}, HyperSim~\cite{roberts2021hypersim}, Mapillary~\cite{neuhold2017mapillary}, Replica~\cite{sucar2021imap}, MVS-Synth~\cite{DeepMVS}, Virtual KITTI~\cite{cabon2020virtual}, Aria Synthetic Environments, and Aria Digital Twin~\cite{pan2023aria}. These datasets, which represent a subset of those in VGGT~\citep{wang2025vggt}, are weighted according to their relative sizes to ensure that each dataset contributes proportionally to the overall training process. Please refer to the Supplemental Material for more training details.

\subsection{Pose Accuracy Benchmark}
\label{sec:exp_pose_acc_bench}
We follow~\citep{maggio2025vggt,elflein2025light3r,brachmann2024acezero} to evaluate camera pose accuracy on three benchmarks: Tanks \& Temples~\citep{knapitsch2017tanks}, TUM-RGBD~\citep{sturm2012benchmark}, and 7-Scenes~\cite{shotton2013scene}. These datasets cover indoor and outdoor environments with image sets and video sequences ranging from $300$ to $20$k images per scene. We exclude comparisons with methods such as~\citep{wang2025vggt, yang2025fast3r, murai2025mast3r, wang2024dust3r} that cannot operate on datasets of this size within a reasonable resource budget. We compute our neural scene representation with $300$ tokens from each anchor frame, with a downsample ratio $r\approx0.2$. All runtime comparisons are conducted under similar GPU throughput conditions, using the Nvidia V100 as the reference hardware.

\noindent
\textbf{Tanks \& Temples} is a dataset includes $21$ large-scale indoor and outdoor scenes, with $150$--$1,100$ images per scene. We follow~\citep{duisterhof2024mast3r, elflein2025light3r} to compare our method with optimization-based (OPT) and feedforward-based (FFD) approaches, defined by whether they utilize an explicit optimization on the 3D structure and camera poses.
In the OPT category, we compare against DF-SfM~\citep{he2024detector}, GLOMAP~\citep{pan2024glomap}, PixelSfM~\citep{lindenberger2021pixel}, VGGSfM~\citep{wang2024vggsfm}, ACE-Zero~\citep{brachmann2024acezero}, FlowMap~\citep{smith2024flowmap}, and MASt3R-SfM~\citep{duisterhof2024mast3r}. In the FFD category, we compare with Spann3R~\citep{wang20253d}, Cut3R~\citep{wang2025continuous}, SLAM3R~\citep{liu2025slam3r}, and Light3R-SfM~\citep{elflein2025light3r}.
We follow~\citep{elflein2025light3r, murai2025mast3r} to report the proportion of camera pairs with relative rotation error ($\text{RRA}@5$) and relative translation error ($\text{RTA}@5$) below $5^o$. 
The mean value of average translation error (ATE) is calculated between the estimated camera poses and the normalized ground truth poses after Procrustes alignment~\cite{Grupp2017evo}. 
As shown in Tab.~\ref{tab:tnt_pose}, our method (FFD), significantly outperforms all feedforward-based baselines, including Cut3R~\cite{wang2025continuous}, Spann3R~\cite{wang20253d}, and SLAM3R~\citep{liu2025slam3r}. 
These methods suffer from pose drift due to their incremental reconstruction manner on sequential input. 
VGGT-SLAM~\citep{maggio2025vggt} achieves the second best results in average, but only recovers keyframe camera poses and fails to reconstruct \textit{Church}, \textit{Courtroom} and \textit{Palace} due to unstable numerical optimization in the $SL(4)$ manifold.
Light3R-SfM~\citep{elflein2025light3r} is more robust by minimizing pose error on the spanning-tree based scene graph, but with much lower accuracy. 
Our method achieves SOTA performance while incurring only a marginal computational overhead compared to Light3R-SfM.
Furthermore, ours (OPT), which utilized 10K iterations of post-refinement, delivers superior performance in all metrics and is competitive with GLOMAP~\citep{pan2024glomap}, with only a marginal increase in runtime.

\noindent
\textbf{TUM-RGBD} is a widely used SLAM benchmark, where each video sequence contains 500--3,000 images. We uniformly select $50$ frames as anchor images. Note that our method is an offline SfM method. Here, we show that our method achieves performance similar to that of some SOTA SLAM systems following the standard split in~\citep{teed2021droid,deng2021adaptive}.
We exclude Cut3R~\citep{wang2025continuous} and SLAM3R~\citep{liu2025slam3r} from the evaluation because they fail on this benchmark.
We evaluate the root mean square error (RMSE) of the absolute trajectory error (ATE) using the \texttt{evo} toolkit~\citep{Grupp2017evo}. As shown in Tab.~\ref{tab:tum_ate}, our method achieves the best results in the uncalibrated setting, without any post-optimization. Results for the calibrated setting are provided in the Supplementary Files. 
In particular, compared to VGGT-SLAM~\citep{maggio2025vggt}, which employs nonlinear factor graph optimization to fuse multiple submaps, our approach achieves higher accuracy without optimization, demonstrating the effectiveness of augmenting scene regression by localization.

\input{tables/table_7scene_pose}

\noindent
\textbf{7 Scenes} is a widely used localization benchmark that provides training and testing split with 2,000--12,000 images per scene. We follow ACE0~\cite{brachmann2024acezero} to evaluate the localization accuracy. The visual localization methods~\cite{brachmann2023accelerated,brachmann2024acezero} will train a scene specific localization network with $1,000$-$4,000$ training images, which takes about 10 minutes to 2 hours depending on whether the ground truth camera poses are given. We report the total time of per scene training and localizing all images.
We uniformly sample 50 images in the testing split as anchor frames and localize all images in the same split. As shown in Tab.~\ref{tab:sevenscenes_reloc_only}, ACE+COLMAP achieves the best performance since it uses ground truth camera poses in training. Without knowing ground truth camera poses, ACE0 and our method achieve the same average localization accuracy. However, ACE0 takes 2 hours to optimize a scene with $4,000$ frames, while \ours uses only 8 minutes. We provide more results on 7-Scenes in Supplemantary Files.

\input{tables/table_tnt_image_set}

\subsection{Novel View Synthesis Benchmark}
As observed in ACE0~\cite{brachmann2024acezero}, the evaluation of camera poses is sometimes unreliable, as the pseudo ground-truth from COLMAP is only an estimation. Thus, we follow ACE0~\cite{brachmann2024acezero} to further evaluate camera pose quality through novel view synthesis.
Specifically, for each method, we first estimate camera poses for all images of a scene. We then split these images into training and testing sets, and train a Nerfacto~\cite{nerfstudio} model on the training set and render images in the testing view.
The rendered images at the testing views are then compared with the ground truth testing images using the Peak Signal-to-Noise Ratio (PSNR) as an indicator of pose accuracy. This evaluation is carried out on the  Mip-NeRF 360~\cite{barron2022mip} and Tanks \& Temples~\cite{knapitsch2017tanks} datasets.
For this evaluation, we enable the post-refinement mentioned in Sec~\ref{sec:refine}, since even slight pose noise can prevent the Nefacto model from converging, resulting in poor PSNR. Note that \textbf{ALL} baselines in this benchmark are optimization-based methods.
Similarly to Sec~\ref{sec:exp_pose_acc_bench}, we exclude comparisons with methods~\citep{wang2025vggt, yang2025fast3r, murai2025mast3r, wang2024dust3r} that cannot operate on datasets of this size within a reasonable resource budget. We exclude~\cite{liu2025slam3r, wang2024spann3r} due to poor performance.

\noindent
\textbf{Tanks \& Temples} has two sub-datasets: images and video sequences. For the image set, each scene contains $150$--$600$ images. The video sequence contains $4,000$--$20,000$ frames. We use $100$ anchor frames at each scene and evaluate our method in both subsets following ACE0~\cite{brachmann2024acezero}. We report the results on the image set and leave the comparison on the video sequences in the Supplemantary. Results of compared methods are quoted from ACE0~\cite{brachmann2024acezero}.

We use COLMAP with the \emph{default} setting CMP (D) as a reference in Tab.~\ref{tab:t2_image_set_psnr}. We enable post-refinement with $10$K iterations.
Our approach achieves the highest PSNR among all baselines, achieving COLMAP-level accuracy while recovering all camera poses in 3-4 minutes. This is significantly faster than COLMAP and ACE0, and is comparable to SLAM systems such as DROID-SLAM, which suffers from pose drifting and produces the lowest PSNR.

\input{tables/table_mip360_psnr}
\noindent
\textbf{Mip-NeRF 360.} 
Mip-NeRF 360~\cite{barron2022mip} is a small-scale dataset containing indoor and outdoor scenes with around $150$--$500$ images per scene. We select $50$ anchor images per scene and enable post-refinement with $10$K iterations. 
The results are reported in Tab.~\ref{tab:mip_psnr_only}. Again, results of all baseline methods are quoted from ACE0~\cite{brachmann2024acezero}. Our approach surpasses all baselines, matching PSNR scores with pseudo ground-truth poses from COLMAP. NoPe-NeRF\cite{bian2022nopenerf} and DROID-SLAM~\cite{teed2021droid} struggle due to wide baselines between images, with NoPe-NeRF\cite{bian2022nopenerf} needing two days of training. BARF~\cite{lin2021barf} has difficulty starting from scratch. ACE0~\cite{brachmann2024acezero} is inferior to \ours in both accuracy and runtime.

\subsection{Ablations}

\input{tables/table_co3d_pose_single}

\noindent
\textbf{Number of Anchor Images.} We evaluate the effect of the number of anchor images on the CO3Dv2 dataset~\cite{reizenstein21co3d}. For each 10 input images, we randomly select $N \in {2, 5, 8, 10}$ anchor images to compute the neural scene representation and localize all $10$ images.
As shown in Tab.~\ref{tab:co3dv2_pose}, our method maintains pose accuracy close to the original VGGT. Its performance drops slowly as the number of anchor images decreases. 
Remarkably, even with as few as two anchor images, our method still delivers strong performance, underscoring its robustness to sparse anchor images.

\begin{figure}
    \centering
    \includegraphics[width=0.95\linewidth]{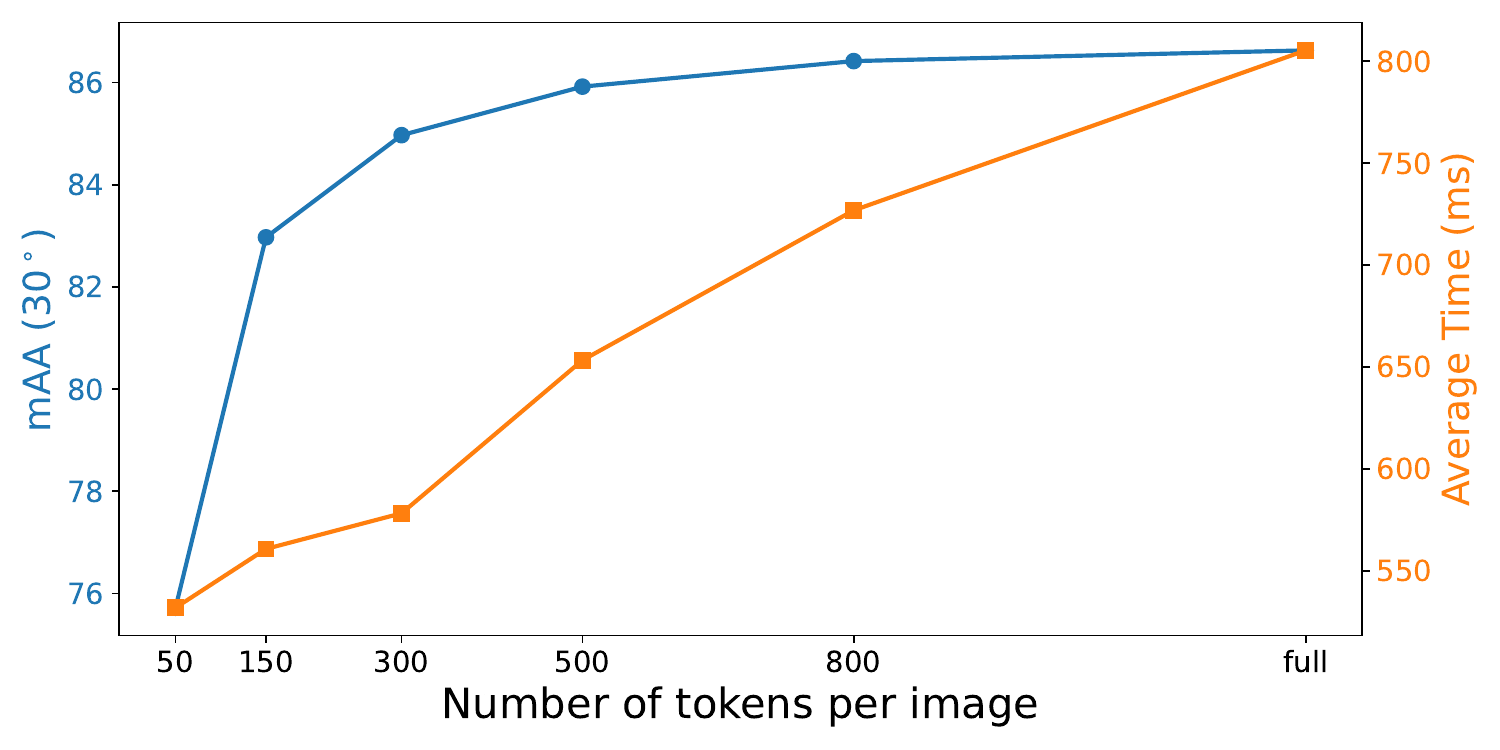}
    \caption{\textbf{Pose Accuracy \& Runtime vs. Tokens per Image.} We choose 300 tokens per image to balance accuracy and efficiency.}
    \label{fig:abla_rank}
    \vspace{-1mm}
\end{figure}
\noindent
\textbf{Number of Tokens/Downsample Ratio $r$.} We investigate the impact of the number of downsampled tokens per anchor image on both pose accuracy and runtime with the number of anchor images fixed at $N=5$, as shown in Fig.~\ref{fig:abla_rank}. Increasing the number of tokens improves pose accuracy, but also leads to a steady growth in processing time. We selected 300 tokens per image as a trade-off, since it achieves reasonable accuracy with low computation cost.

\input{tables/table_abla_train_stra}
\noindent
\textbf{Training Strategy.} We further evaluate the effect of our training strategy. Specifically, during training, our method selects a random number of tokens per image. We compare it with two alternatives: (i) using a fixed number of tokens (300 per image) and (ii) applying average pooling over image tokens to achieve a $4\times$ downsampling (resulting in approximately 340 tokens per image). As shown in Tab.~\ref{tab:abla_train_stra}, our variant token strategy yields more accurate pose estimation. In particular, our method outperforms the average pooling baseline. We attribute it to dropout~\cite{srivastava2014dropout}, where our random selection acts as a regularization mechanism, which improves generalization. Moreover, the variant token strategy offers greater flexibility than pooling, enabling an explicit trade-off between accuracy and efficiency.

%% file: tables/table_tnt_pose.tex
\newcommand{\bgcolor}[2]{\setlength{\fboxsep}{0pt}\colorbox{#1}{\strut #2}}
\newcommand{\BGcolor}[3][HTML]{\definecolor{mycolor}{HTML}{#2}\bgcolor{mycolor}{#3}}

\begin{table}[!tp]
\begin{center}
\adjustbox{valign=t,width=\linewidth}{
\sisetup{detect-all=true,detect-weight=true}
\begin{tabular}{llccccc}
\toprule
 Method  & Align. & RRA@5 $\uparrow$ & RTA@5 $\uparrow$ & ATE $\downarrow$ & Reg. $\uparrow$ & Time [s] $\downarrow$ \\
\midrule

 COLMAP~\citep{schonberger2016structure} & OPT & {\cellcolor[HTML]{FFFFFF}} \color[HTML]{000000} GT & {\cellcolor[HTML]{FFFFFF}} \color[HTML]{000000} GT & {\cellcolor[HTML]{FFFFFF}} \color[HTML]{000000} GT & {\cellcolor[HTML]{FFFFFF}} \color[HTML]{000000} GT & {\cellcolor[HTML]{FFFFFF}} \color[HTML]{000000} - \\
 GLOMAP~\citep{pan2024glomap} & OPT & \firstc \color[HTML]{000000} 75.8 & \secondc \color[HTML]{000000} 76.7 & \secondc \color[HTML]{000000} 0.010 & \firstc \color[HTML]{000000} 100.0 & \thirdc \color[HTML]{000000} 1977 \\
 ACE0~\citep{brachmann2024acezero} & OPT &  \color[HTML]{000000} 56.9 &  \color[HTML]{000000} 57.9 &  \color[HTML]{000000} 0.015 & \firstc \color[HTML]{000000} 100.0 &  \color[HTML]{000000} 5499 \\
 DF-SfM~\citep{he2024detector} & OPT & \thirdc \color[HTML]{000000} 69.6 & \thirdc \color[HTML]{000000} 69.3 &  \color[HTML]{000000} 0.014 & {\cellcolor[HTML]{E4E7BE}} \color[HTML]{000000} 76.2 & {\cellcolor[HTML]{FFFFFF}} \color[HTML]{000000} - \\
 FlowMap~\citep{smith2024flowmap} & OPT &  \color[HTML]{000000} 31.7 &  \color[HTML]{000000} 35.7 &  \color[HTML]{000000} 0.017 & \thirdc \color[HTML]{000000} 66.7 & {\cellcolor[HTML]{FFFFFF}} \color[HTML]{000000} - \\
 VGGSfM~\citep{wang2024vggsfm} & OPT & {\cellcolor[HTML]{FFFFFF}} \color[HTML]{000000} - & {\cellcolor[HTML]{FFFFFF}} \color[HTML]{000000} - & {\cellcolor[HTML]{FFFFFF}} \color[HTML]{000000} - &  \color[HTML]{000000} 0.0 &  \color[HTML]{000000} 2134 \\
 MASt3R-SfM~\citep{duisterhof2024mast3r} & OPT &  \color[HTML]{000000} 49.2 &  \color[HTML]{000000} 54.0 &  \thirdc\color[HTML]{000000} 0.011 & \firstc \color[HTML]{000000} 100.0 &  \color[HTML]{000000} 2723 \\
 DROID-SLAM~\citep{teed2021droid} & OPT &  \color[HTML]{000000} 31.3 &  \color[HTML]{000000} 40.3 &  \thirdc\color[HTML]{000000} 0.021 & \firstc \color[HTML]{000000} 100.0 & \secondc \color[HTML]{000000} 240 \\
 \ours-OPT & OPT & \secondc \color[HTML]{000000} 71.5 & \firstc \color[HTML]{000000} 77.7 & \firstc \color[HTML]{000000} 0.008 & \firstc \color[HTML]{000000} 100.0 & \firstc \color[HTML]{000000} 233 \\
\cmidrule{1-7}
  Cut3R$^\dagger$~\citep{wang2025continuous} & FFD &  \color[HTML]{000000} 18.8 &  \color[HTML]{000000} 25.8 &  \color[HTML]{000000} 0.017 & \firstc \color[HTML]{000000} 100.0 & \firstc \color[HTML]{000000} 42 \\
 Spann3R$^\dagger$~\citep{wang20253d} & FFD &  \color[HTML]{000000} 22.1 &  \color[HTML]{000000} 30.7 &  \color[HTML]{000000} 0.016 & \firstc \color[HTML]{000000} 100.0 & \color[HTML]{000000} 116 \\
 SLAM3R$^\dagger$~\citep{liu2025slam3r} & FFD &  \color[HTML]{000000} 20.3 &  \color[HTML]{000000} 24.7 &  \thirdc\color[HTML]{000000} 0.015 & \firstc \color[HTML]{000000} 100.0 &  \thirdc 70 \\
 Light3R-SfM~\citep{elflein2025light3r} & FFD &  \thirdc\color[HTML]{000000} 52.0 & \thirdc \color[HTML]{000000} 52.8 &  \secondc\color[HTML]{000000} 0.011 & \firstc \color[HTML]{000000} 100.0 & \secondc \color[HTML]{000000} 63 \\
 VGGT-SLAM*$^\triangle$~\citep{maggio2025vggt} & FFD &  \secondc\color[HTML]{000000} 57.3 &  \secondc\color[HTML]{000000} 67.9 &  \firstc\color[HTML]{000000} 0.008 & \firstc \color[HTML]{000000} 100.0 &  \color[HTML]{000000} 238 \\
  \ours & FFD & \firstc \color[HTML]{000000} 70.4 & \firstc \color[HTML]{000000} 74.7 & \firstc \color[HTML]{000000} 0.008 & \firstc \color[HTML]{000000} 100.0 &  \color[HTML]{000000} 81 \\
\bottomrule
\end{tabular}
}
\caption{\textbf{Pose Estimation on Tanks \& Temples.~\citep{knapitsch2017tanks}.} This dataset contains on average over 300 images per scene. We visualize the results using three colors: \TCBTableBest {Best}, \TCBTableSecond{Second}, and \TCBTableThird{Third}. A `–' indicates cases where all scenes failed to converge or the running time is unavailable; $^\dagger$ denotes that sequential input is required; * indicates evaluation on keyframes; and $\triangle$ denotes that some sequences fail. ‘OPT’ stands for optimization-based and ‘FFD’ stands for feedforward-based.}
\label{tab:tnt_pose}
\end{center}
\vspace{-8mm}
\end{table}

%% file: tables/table_tum.tex
\begin{table*}[h]
    \vspace{-0.5em}
    \centering
    
    \scriptsize
    \begin{tabular}{l|cccccccccc} 
    \toprule
    \multirow{2}{*}{Method} & \multicolumn{9}{c}{Sequence} &  \multirow{2}{*}{Avg.}  \\ 
    \cmidrule(lr){2-10}
    &\texttt{360} &\texttt{desk} &\texttt{desk2} &\texttt{floor} &\texttt{plant} &\texttt{room } &\texttt{rpy} &\texttt{teddy} &\texttt{xyz}  \\
    \midrule
    DROID-SLAM*~\cite{teed2021droid} &0.202 &  \thirdc 0.032 &0.091 & \secondc 0.064 &0.045 &0.918 &0.056 &0.045 & \firstc 0.012 &0.158 \\
    \mr-SLAM*~\cite{murai2025mast3r} &\firstc {0.070} & 0.035 & \thirdc 0.055 & \firstc 0.056 &0.035 & 0.118 & \thirdc0.041 &0.114 & 0.020 & \thirdc 0.060 \\
    VGGT-SLAM~(\Simthree)~\citep{maggio2025vggt} & \thirdc 0.123 & 0.040 & \thirdc 0.055 & 0.254 & \firstc 0.022 & \firstc 0.088 &\thirdc 0.041 & \firstc 0.032 &\thirdc 0.016 &  0.074 \\
    VGGT-SLAM~(\SLfour)~\citep{maggio2025vggt} & \secondc 0.071 & \secondc 0.025 & \firstc 0.040 & 0.141 & \secondc 0.023 & \secondc 0.102 & \secondc 0.030 & \secondc 0.034 & \secondc 0.014 & \secondc 0.053 \\
    \ours(Offline)  & \firstc0.070 & \firstc \firstc0.024 & \secondc 0.042 & \thirdc 0.107 & \thirdc 0.031 & \thirdc 0.113 & \firstc 0.020 &  \thirdc0.037 & \firstc 0.012 & \firstc 0.051 \\
    \bottomrule
    \end{tabular}
    \caption{\textbf{Root Mean Square Error~(RMSE) of Absolute Trajectory Error~(ATE) on the TUM RGB-D~\citep{sturm2012benchmark} dataset (unit: m)}.
    We evaluate methods under an uncalibrated configration following VGGT-SLAM~\cite{maggio2025vggt}, while methods marked with * indicate the intrinsic matrics are provided by GeoCalib~\cite{veicht2024geocalib}.
    We color result in: \TCBTableBest {Best}, \TCBTableSecond{Second}, and \TCBTableThird{Third}.}
    \label{tab:tum_ate}
    \vspace{-5mm}
\end{table*}

%% file: tables/table_7scene_pose.tex
\begin{table}[!tp]
\begin{center}
\adjustbox{valign=t,width=\linewidth}{
\sisetup{detect-all=true,detect-weight=true}
\begin{tabular}{c|cccc}
\hhline{-----}
\hhline{-----}

& \multicolumn{1}{@{}c@{}}{~ACE~}
& \multicolumn{1}{@{}c@{}}{~ACE~}
& \multicolumn{1}{@{}c@{}}{~ACE0~}%
& \multicolumn{1}{@{}c@{}}{~Ours~} \\

Prior
& \multicolumn{1}{@{}c@{}}{~KinectFusion~}
& \multicolumn{1}{@{}c@{}}{~COLMAP~}
& \multicolumn{1}{@{}c@{}}{~OPT.~}%
& \multicolumn{1}{@{}c@{}}{~FFD.~} \\

\hhline{-----}
Chess       & 96.0\% & 100.0\% & 100.0\% & 98.8\%\\
Fire        & 98.4\% & 99.5\%  & 98.8\%  & 100.0\%  \\
Heads       & 100.0\%& 100.0\% & 100.0\% & 100.0\% \\
Office      & 36.9\% & 100.0\% & 99.1\%  & 87.4\%  \\
Pumpkin     & 47.3\% & 100.0\% & 99.9\%  & 92.8\% \\
Redkitchen  & 47.8\% & 98.9\%  & 98.1\%  & 89.9\%  \\
Stairs      & 74.1\% & 85.0\%  & 61.0\%  & 87.9\% \\
\hhline{-----}
Average     & 74.1\% & 97.6\%  & 93.8\%  & 93.8\%  \\
Average Time   & 14min & 14min  & 2h  & 8min   \\
\hhline{-----}
\end{tabular}
}
 \caption{%
            \textbf{Localization on 7-Scenes.} Percentage of pose error under \mbox{(5cm, 5$^\circ$)}, compared to pseudo ground truth computed by COLMAP. ACE requires known camera poses during training. Our method achieves comparable localization accuracy to ACE0, where neither approach relies on camera poses in the training set. However, our method is significantly faster than ACE0, which performs self-supervised optimization.
        }\label{tab:sevenscenes_reloc_only}
    \vspace{-7mm}
\label{tab:7scens_reloc_mip360_psnr}
\end{center}
\end{table}

%% file: tables/table_tnt_image_set.tex
\begin{table}[t]%
    \centering%
    \setlength{\tabcolsep}{1pt}%
    \footnotesize%
    \adjustbox{valign=t,width=\linewidth}{%
    \begin{tabular}{clccccccccccccccccc}
        \toprule
         ~ &%
         ~ &%
        \hspace{-6pt}\parbox[t]{0mm}{\multirow{3}{*}{\rotatebox[origin=c]{90}{Frames}}} &%
        \multicolumn{3}{c|}{~} &%
         & ~ & &%
        \multicolumn{3}{c}{DROID-} &%
        \multicolumn{3}{c}{~} &
        \multicolumn{3}{c}{~} %
        \\
         ~ &%
         ~ &%
         ~ &%
        \multicolumn{3}{c|}{CMP} &%
        \multicolumn{3}{c}{Reality} &%
        \multicolumn{3}{c}{SLAM$^\dagger$} &%
        \multicolumn{3}{c}{ACE0~\citep{brachmann2024acezero}} &
        \multicolumn{3}{c}{Ours} %
        \\
         ~ &%
         ~ &%
         ~ &%
        \multicolumn{3}{c|}{(D)}  &%
        \multicolumn{3}{c}{Capture} &%
        \multicolumn{3}{c}{\cite{teed2021droid}} &%
        \multicolumn{3}{c}{~~~~} &
        \multicolumn{3}{c}{~~~~} %
        \\
        \midrule
        \parbox[t]{3mm}{
        \multirow{8}{*}{\rotatebox[origin=c]{90}{Training}}} 
        & Barn & 410 & 
        \multicolumn{3}{c|}{\begin{tabular}{ccc} & ~{24.0}~ & \end{tabular}} &  
         \multicolumn{3}{c}{\begin{tabular}{ccc} & ~{\secondc21.2}~ & \end{tabular}} &  
         \multicolumn{3}{c}{\begin{tabular}{ccc} & ~{\thirdc19.0}~ & \end{tabular}} &  
         \multicolumn{3}{c}{\begin{tabular}{ccc} & ~{16.5}~ & \end{tabular}} &
         \multicolumn{3}{c}{\begin{tabular}{ccc} &~{\firstc23.5}~ & \end{tabular}}\\

        &&&&&&&&&&&&&&&&&&\\[-10pt]
         & Catpr. & 383 &  
         \multicolumn{3}{c|}{\begin{tabular}{ccc} & ~{17.1}~ & \end{tabular}} &  
         \multicolumn{3}{c}{\begin{tabular}{ccc} & ~{15.9}~ & \end{tabular}} &  
         \multicolumn{3}{c}{\begin{tabular}{ccc} & ~{\thirdc16.6}~ & \end{tabular}} &  
         \multicolumn{3}{c}{\begin{tabular}{ccc} & ~{\firstc16.9}~ & \end{tabular}} &
         \multicolumn{3}{c}{\begin{tabular}{ccc}&~{\secondc16.8}~ & \end{tabular}}  \\
        &&&&&&&&&&&&&&&&&&\\[-10pt]
         & Church & 507 & 
        \multicolumn{3}{c|}{\begin{tabular}{ccc} & ~{18.3}~ & \end{tabular}} &  
         \multicolumn{3}{c}{\begin{tabular}{ccc} & ~{\firstc17.6}~ & \end{tabular}} &  
         \multicolumn{3}{c}{\begin{tabular}{ccc} & ~{14.3}~ & \end{tabular}} &  
         \multicolumn{3}{c}{\begin{tabular}{ccc} & ~{\secondc17.2}~ & \end{tabular}} &
         \multicolumn{3}{c}{\begin{tabular}{ccc} &~{\thirdc17.0}~ & \end{tabular}}\\
         &&&&&&&&&&&&&&&&&&\\[-10pt]
         & Ignatius & 264 & 
         \multicolumn{3}{c|}{\begin{tabular}{ccc} & ~{20.1}~ & \end{tabular}} &  
         \multicolumn{3}{c}{\begin{tabular}{ccc} & ~{17.7}~ & \end{tabular}} &  
         \multicolumn{3}{c}{\begin{tabular}{ccc} & ~{\thirdc17.8}~ & \end{tabular}} &  
         \multicolumn{3}{c}{\begin{tabular}{ccc} & ~{\firstc19.8}~ & \end{tabular}} &
         \multicolumn{3}{c}{\begin{tabular}{ccc} &~{\secondc19.5}~ & \end{tabular}}\\
          &&&&&&&&&&&&&&&&&&\\[-10pt]
         & MtgRm. & 371 &
          \multicolumn{3}{c|}{\begin{tabular}{ccc} & ~{18.6}~ & \end{tabular}} &
         \multicolumn{3}{c}{\begin{tabular}{ccc} & ~{\secondc18.1}~ & \end{tabular}} &  
         \multicolumn{3}{c}{\begin{tabular}{ccc} & ~{15.6}~ & \end{tabular}} &  
         \multicolumn{3}{c}{\begin{tabular}{ccc} & ~{\thirdc18.0}~ & \end{tabular}} &
         \multicolumn{3}{c}{\begin{tabular}{ccc} &~{\firstc19.5}~ & \end{tabular}}  \\

         &&&&&&&&&&&&&&&&&&\\[-10pt]
         & Truck & 251 &
         \multicolumn{3}{c|}{\begin{tabular}{ccc} & ~{21.1}~ & \end{tabular}} & 
         \multicolumn{3}{c}{\begin{tabular}{ccc} & ~{\thirdc19.0}~ & \end{tabular}} &  
         \multicolumn{3}{c}{\begin{tabular}{ccc} & ~{18.3}~ & \end{tabular}} &  
         \multicolumn{3}{c}{\begin{tabular}{ccc} & ~{\secondc20.1}~ & \end{tabular}} & 
         \multicolumn{3}{c}{\begin{tabular}{ccc} &~{\firstc20.9}~ & \end{tabular}}\\

        &&&&&&&&&&&&&&&&&&\\[-10pt]
        \hhline{~------------------}
        &&&&&&&&&&&&&&&&&&\\[-10pt]
         & Average & 364 &
         \multicolumn{3}{c|}{\begin{tabular}{ccc} & ~{19.9}~ & \end{tabular}} &  
         \multicolumn{3}{c}{\begin{tabular}{ccc} & ~{\secondc18.2}~ & \end{tabular}} &  
         \multicolumn{3}{c}{\begin{tabular}{ccc} & ~{16.9}~ & \end{tabular}} &  
         \multicolumn{3}{c}{\begin{tabular}{ccc} & ~{\thirdc18.1}~ & \end{tabular}} & 
         \multicolumn{3}{c}{\begin{tabular}{ccc} &~{\firstc19.5}~ & \end{tabular}}\\

        &&&&&&&&&&&&&&&&&&\\[-10pt]
        \hline
         & Time &      & 
         \multicolumn{3}{c|}{\begin{tabular}{ccc} & ~1h~ & \end{tabular}} &
         \multicolumn{3}{c}{\begin{tabular}{ccc} & ~3min~ & \end{tabular}} &  
         \multicolumn{3}{c}{\begin{tabular}{ccc} & ~5min~ & \end{tabular}} &  
         \multicolumn{3}{c}{\begin{tabular}{ccc} & ~1.1h~ & \end{tabular}} &  
         \multicolumn{3}{c}{\begin{tabular}{ccc} & ~3.5min~ & \end{tabular}} \\

        \hline
         &&&&&&&&&&&&&&&&&&\\[-5pt]
        \hline

        &&&&&&&&&&&&&&&&&& \\[-10pt]
        \parbox[t]{5mm}{\multirow{8}{*}{\rotatebox[origin=c]{90}{Intermediate}}} 
        & Family & 152 & 
         \multicolumn{3}{c|}{\begin{tabular}{ccc} & ~{19.5}~ & \end{tabular}} & 
        \multicolumn{3}{c}{\begin{tabular}{ccc} & ~{\thirdc18.8}~ & \end{tabular}} & 
        \multicolumn{3}{c}{\begin{tabular}{ccc} & ~{17.6}~ & \end{tabular}} &  
        \multicolumn{3}{c}{\begin{tabular}{ccc} & ~{\secondc19.0}~ & \end{tabular}} &
        \multicolumn{3}{c}{\begin{tabular}{ccc} &~{\firstc20.6}~ & \end{tabular}}\\
        
        &&&&&&&&&&&&&&&&&& \\[-10pt]
         & Francis & 302 &
         \multicolumn{3}{c|}{\begin{tabular}{ccc} & ~{21.6}~ & \end{tabular}} &    
         \multicolumn{3}{c}{\begin{tabular}{ccc} & ~{\secondc20.7}~ & \end{tabular}} &  
         \multicolumn{3}{c}{\begin{tabular}{ccc} & ~{\secondc20.7}~ & \end{tabular}} &  
         \multicolumn{3}{c}{\begin{tabular}{ccc} & ~{\thirdc20.1}~ & \end{tabular}} &
         \multicolumn{3}{c}{\begin{tabular}{ccc} &~{\firstc21.8}~ & \end{tabular}} \\

         &&&&&&&&&&&&&&&&&&\\[-10pt]
         & Horse & 151 & 
         \multicolumn{3}{c|}{\begin{tabular}{ccc} & ~{19.2}~ & \end{tabular}} & 
         \multicolumn{3}{c}{\begin{tabular}{ccc} & ~{\thirdc19.0}~ & \end{tabular}} &  
         \multicolumn{3}{c}{\begin{tabular}{ccc} & ~{16.3}~ & \end{tabular}} &  
         \multicolumn{3}{c}{\begin{tabular}{ccc} & ~{\secondc19.5}~ & \end{tabular}} &
         \multicolumn{3}{c}{\begin{tabular}{ccc} &~{\firstc20.1}~ & \end{tabular}} \\

         &&&&&&&&&&&&&&&&&&\\[-10pt]
         & LightH. & 309 & 
         \multicolumn{3}{c|}{\begin{tabular}{ccc} & ~{16.6}~ & \end{tabular}} &   
         \multicolumn{3}{c}{\begin{tabular}{ccc} & ~{\thirdc16.5}~ & \end{tabular}} & 
         \multicolumn{3}{c}{\begin{tabular}{ccc} & ~{13.6}~ & \end{tabular}} &  
         \multicolumn{3}{c}{\begin{tabular}{ccc} & ~{\secondc17.5}~ & \end{tabular}} &
         \multicolumn{3}{c}{\begin{tabular}{ccc} &~{\firstc18.2}~ & \end{tabular}} \\

         &&&&&&&&&&&&&&&&&&\\[-10pt]
         & PlayGd. & 307 & 
         \multicolumn{3}{c|}{\begin{tabular}{ccc} & ~{19.1}~ & \end{tabular}} &  
         \multicolumn{3}{c}{\begin{tabular}{ccc} & ~{\secondc19.2}~ & \end{tabular}} &  
         \multicolumn{3}{c}{\begin{tabular}{ccc} & ~{11.4}~ & \end{tabular}} &  
         \multicolumn{3}{c}{\begin{tabular}{ccc} & ~{\thirdc18.7}~ & \end{tabular}} & 
         \multicolumn{3}{c}{\begin{tabular}{ccc} &~{\firstc20.3}~ & \end{tabular}}\\

        &&&&&&&&&&&&&&&&&&\\[-10pt]
         & Train & 301 & 
         \multicolumn{3}{c|}{\begin{tabular}{ccc} & ~{16.8}~ & \end{tabular}} & 
         \multicolumn{3}{c}{\begin{tabular}{ccc} & ~{\secondc15.4}~ & \end{tabular}} &  
         \multicolumn{3}{c}{\begin{tabular}{ccc} & ~{\thirdc13.8}~ & \end{tabular}} &  
         \multicolumn{3}{c}{\begin{tabular}{ccc} & ~{\firstc16.2}~ & \end{tabular}} & 
         \multicolumn{3}{c}{\begin{tabular}{ccc} &~{\firstc16.2}~ & \end{tabular}}  \\

         &&&&&&&&&&&&&&&&&&\\[-10pt]
          \hhline{~------------------}
        &&&&&&&&&&&&&&&&&&\\[-10pt]
         & Average & 254 & 
         \multicolumn{3}{c|}{\begin{tabular}{ccc} & ~{18.8}~ & \end{tabular}} &  
         \multicolumn{3}{c}{\begin{tabular}{ccc} & ~{\thirdc18.3}~ & \end{tabular}} &  
         \multicolumn{3}{c}{\begin{tabular}{ccc} & ~{15.6}~ & \end{tabular}} &  
         \multicolumn{3}{c}{\begin{tabular}{ccc} & ~{\secondc18.5}~ & \end{tabular}} & 
         \multicolumn{3}{c}{\begin{tabular}{ccc} &~{\firstc19.5}~ & \end{tabular}}\\

         &&&&&&&&&&&&&&&&&&\\[-10pt]
        \hline
         & Time &      & 
         \multicolumn{3}{c|}{\begin{tabular}{ccc} & ~32min~ & \end{tabular}} &  
         \multicolumn{3}{c}{\begin{tabular}{ccc} & ~2min~ & \end{tabular}} &  
         \multicolumn{3}{c}{\begin{tabular}{ccc} & ~3min~ & \end{tabular}} &  
         \multicolumn{3}{c}{\begin{tabular}{ccc} & ~1.3h~ & \end{tabular}} &   
         \multicolumn{3}{c}{\begin{tabular}{ccc} & ~3min~ & \end{tabular}} \\
        \hline
        &&&&&&&&&&&&&&&&&&\\[-5pt]
        \hline
        
        &&&&&&&&&&&&&&&&&&\\[-10pt]
    \parbox[t]{5mm}{\multirow{7}{*}{\rotatebox[origin=c]{90}{Advanced}}} & 
    Audtrm. & 302 & 
    \multicolumn{3}{c|}{\begin{tabular}{ccc} & ~{19.6}~ & \end{tabular}} & 
    \multicolumn{3}{c}{\begin{tabular}{ccc} & ~{12.2}~ & \end{tabular}} & 
    \multicolumn{3}{c}{\begin{tabular}{ccc} & ~{\thirdc16.7}~ & \end{tabular}} & 
    \multicolumn{3}{c}{\begin{tabular}{ccc} & ~{\secondc18.7}~ & \end{tabular}} & 
    \multicolumn{3}{c}{\begin{tabular}{ccc} &~{\firstc20.3}~ & \end{tabular}}\\

     &&&&&&&&&&&&&&&&&& \\[-10pt]
     & BallRm. & 324 & 
     \multicolumn{3}{c|}{\begin{tabular}{ccc} & ~{16.3}~ & \end{tabular}} &  
     \multicolumn{3}{c}{\begin{tabular}{ccc} & ~{\firstc 18.3}~ & \end{tabular}} & 
     \multicolumn{3}{c}{\begin{tabular}{ccc} & ~{13.1}~ & \end{tabular}} &  
     \multicolumn{3}{c}{\begin{tabular}{ccc} & ~{\secondc17.9}~ & \end{tabular}} &
     \multicolumn{3}{c}{\begin{tabular}{ccc} &~{\thirdc 14.8}~ & \end{tabular}} \\

     &&&&&&&&&&&&&&&&&& \\[-10pt]
     & CortRm. & 301 & 
     \multicolumn{3}{c|}{\begin{tabular}{ccc} & ~{18.2}~ & \end{tabular}} &  
     \multicolumn{3}{c}{\begin{tabular}{ccc} & ~{\secondc17.2}~ & \end{tabular}} & 
     \multicolumn{3}{c}{\begin{tabular}{ccc} & ~{12.3}~ & \end{tabular}} & 
     \multicolumn{3}{c}{\begin{tabular}{ccc} & ~{\thirdc17.1}~ & \end{tabular}} & 
     \multicolumn{3}{c}{\begin{tabular}{ccc} &~{\firstc17.4}~ & \end{tabular}} \\

         &&&&&&&&&&&&&&&&&& \\[-10pt]
         & Palace & 509 & 
         \multicolumn{3}{c|}{\begin{tabular}{ccc} & ~{14.2}~ & \end{tabular}} &  
         \multicolumn{3}{c}{\begin{tabular}{ccc} & ~{\secondc11.7}~ & \end{tabular}} &  
         \multicolumn{3}{c}{\begin{tabular}{ccc} & ~{\thirdc10.8}~ & \end{tabular}} &  
         \multicolumn{3}{c}{\begin{tabular}{ccc} & ~{10.7}~ & \end{tabular}} & 
         \multicolumn{3}{c}{\begin{tabular}{ccc} &~{\firstc14.3}~ & \end{tabular}} \\

          &&&&&&&&&&&&&&&&&&\\[-10pt]
         & Temple & 302 & 
         \multicolumn{3}{c|}{\begin{tabular}{ccc} & ~{18.1}~ & \end{tabular}} & 
         \multicolumn{3}{c}{\begin{tabular}{ccc} & ~{\secondc15.7}~ & \end{tabular}} &  
         \multicolumn{3}{c}{\begin{tabular}{ccc} & ~{\thirdc11.8}~ & \end{tabular}} &  
         \multicolumn{3}{c}{\begin{tabular}{ccc} & ~{9.7}~ & \end{tabular}} &
         \multicolumn{3}{c}{\begin{tabular}{ccc} &~{\firstc17.8}~ & \end{tabular}}\\

         &&&&&&&&&&&&&&&&&&\\[-10pt]
        \hhline{~------------------}
        &&&&&&&&&&&&&&&&&&\\[-10pt]
         & Average & 348 & 
        \multicolumn{3}{c|}{\begin{tabular}{ccc} & ~{17.3}~ & \end{tabular}} &  
         \multicolumn{3}{c}{\begin{tabular}{ccc} & ~{\secondc15.0}~ & \end{tabular}} &  
         \multicolumn{3}{c}{\begin{tabular}{ccc} & ~{12.9}~ & \end{tabular}} &  
         \multicolumn{3}{c}{\begin{tabular}{ccc} & ~{\thirdc14.8}~ & \end{tabular}} & 
         \multicolumn{3}{c}{\begin{tabular}{ccc} &~{\firstc16.9}~ & \end{tabular}}\\

        &&&&&&&&&&&&&&&&&&\\[-10pt]
        \hline
         & Time &      & 
         \multicolumn{3}{c|}{\begin{tabular}{ccc} & ~1h~ & \end{tabular}} &  
         \multicolumn{3}{c}{\begin{tabular}{ccc} & ~2min~ & \end{tabular}} &  
         \multicolumn{3}{c}{\begin{tabular}{ccc} & ~4min~ & \end{tabular}} & 
         \multicolumn{3}{c}{\begin{tabular}{ccc} & ~1h~ & \end{tabular}} & 
         \multicolumn{3}{c}{\begin{tabular}{ccc} & ~3.5min~ & \end{tabular}}   \\
        \bottomrule
         
    \end{tabular}
    }
    \caption{\textbf{Tanks \& Temples.}  Pose accuracy via view synthesis with Nerfacto~\cite{nerfstudio}. We report the PSNR in dB and the average reconstruction time. We color code in: \TCBTableBest {Best}, \TCBTableSecond{Second}, and \TCBTableThird{Third}. $^{\dagger}$ indicates methods needing sequential inputs.}%
        \vspace{-2mm}
    \label{tab:t2_image_set_psnr}%
\end{table}

%% file: tables/table_mip360_psnr.tex
\begin{table}[!tp]
\begin{center}
\adjustbox{valign=t,width=\linewidth}{
\sisetup{detect-all=true,detect-weight=true}
\begin{tabular}{cc||ccccc}
\hhline{-------}
\hhline{-------}

&  Pseudo GT
&  DROID-
&  BARF
&  Nope-
&  ACE0
&  Ours\\ 

& \multicolumn{1}{@{}c||@{}}{~COLMAP~} 
& SLAM$^\dagger$\cite{teed2021droid}
& \cite{lin2021barf}
&NeRF\cite{bian2022nopenerf}
&\cite{brachmann2024acezero}\\ 
\hhline{-------}
Bicycle    & 21.5 & 10.9 & 11.9 & \thirdc12.2 & \secondc 18.7 & \firstc 20.50 \\
Bonsai     & 27.6 & 10.9 & 12.5 & \thirdc14.8 &\secondc 25.8 & \firstc 26.76 \\
Counter    & 25.5 & \thirdc12.9 & 11.9 & 11.6 & \secondc24.5 & \firstc 25.51 \\
Garden     & 26.3 & \thirdc16.7 & 13.3 & 13.8 & \firstc 25.0 & \secondc24.92 \\
Kitchen    & 27.4 & 13.9 & 13.3 & \thirdc14.4 & \secondc 26.1 & \firstc 27.43 \\
Room       & 28.0 & 11.3 & 11.9 & \thirdc14.3 & \secondc 19.8 & \firstc 27.46 \\
Stump      & 16.8 & 13.9 & \thirdc15.0 & 13.9 & \secondc 20.5 & \firstc 20.83 \\
\hhline{-------}
Average    & 24.7 & 12.9 & 12.8 & \thirdc13.5 & \secondc22.9 & \firstc 24.77 \\
Average Time  &  1h & 2min & 4h & $\geq$24h & 8h &5min    \\
\hhline{-------}

\end{tabular}
}

 \caption{%
          \textbf{Mip-NeRF 360.} Pose accuracy via view synthesis PSNR. Higher is better.  We color code in: \TCBTableBest {Best}, \TCBTableSecond{Second}, and \TCBTableThird{Third}. $^\dagger$ indicates methods needing sequential inputs.%
        }\label{tab:mip_psnr_only}
            
\label{tab:mip360_psnr}
\end{center}
\vspace{-10mm}
\end{table}

%% file: tables/table_co3d_pose_single.tex
\begin{table}[tbp]

\makeatletter\def\@captype{table}
\centering

    \resizebox{0.95\columnwidth}{!}{
\begin{tabular}{c|c|c|c|c}
\toprule

\multirow{2}{*}{Method}  & Global& \multicolumn{3}{c}{Co3Dv2$\uparrow$} \\ 
\cline{3-5} 
                       & Align. & RRA@15  & RTA@15 & mAA@30 \\ 
\midrule

Colmap~\cite{schoenberger2016sfm}        & OPT & 31.6    & 27.3   & 25.3              \\
Glomap~\cite{pan2024glomap}                & OPT & 45.9    & 40.3   & 37.3     \\
PixSfM~\cite{lindenberger2021pixel}       & OPT & 33.7    & 32.9   & 30.1              \\
VGGSfM~\cite{wang2024vggsfm}               & OPT & 92.1 & 88.3 & 74.0 \\
\dustr-GA~\cite{wang2024dust3r}            & OPT & {96.2}    & 86.8   & 76.7    \\ 
\mastr-SfM~\cite{duisterhof2024mast3r}  & OPT & 96.0 & {93.1}   &  {88.0}   \\
\cline{1-5}
PoseDiff ~\cite{wang2023posediffusion}     & FFD & 80.5    & 79.8   & 66.5    \\
RelPose++~\cite{lin2024relpose++}          & FFD & 82.3    &  77.2      & 65.1   \\
\spannr~\cite{wang20253d}             & FFD & 89.5 & 83.2 & 70.3    \\
\mastr\textbf{*}~\cite{leroy2024grounding}          & FFD &  94.5    &  80.9      & 68.7   \\
\lightr ~\cite{elflein2025light3r}  & FFD & 94.7 & 85.8  &  72.8    \\  %
VGGT~\cite{wang2025vggt}          & FFD &  98.4    &  94.8      & 88.2   \\
\cline{1-5}

\textbf{\ours} $N=10$         & FFD        & 98.3    & 94.0  & 88.1 \\
\textbf{\ours} $N=8$         & FFD        & 98.2    & 93.6  & 87.3 \\
\textbf{\ours} $N=5$         & FFD        & 97.7   & 92.2   &  85.0 \\ 
\textbf{\ours} $N=2$         & FFD        & 96.4 & 89.7 & 78.5 \\

\bottomrule
\end{tabular}
}

\caption{\textbf{Ablation on the number of anchor views for pose estimation performance on CO3Dv2~\cite{reizenstein21co3d}}. We evaluate pose estimation accuracy by varying the number of anchor views from 10 input images, randomly sampled from each sequence.}
    \vspace{-5mm}
\label{tab:co3dv2_pose}
\end{table}

%% file: tables/table_abla_train_stra.tex
\begin{table}[tbp]
\makeatletter\def\@captype{table}
\centering

\renewcommand{\arraystretch}{0.8} 
\scriptsize 

\resizebox{0.95\columnwidth}{!}{
\begin{tabular}{c|c|c|c|c}
\toprule
\multirow{2}{*}{Method}  &  \multicolumn{4}{c}{Co3Dv2$\uparrow$} \\ 
& mAA@30 & mAA@5 & RRA@15  & RTA@15 \\ 
\midrule
\textbf{\ours}        & \textbf{87.3} & \textbf{57.6} & \textbf{98.3}  & \textbf{93.6} \\
Fixed token         & 86.5 & 53.6 & 98.2  & 93.6 \\
Avg. Pooling        & 86.5 & 53.5 & 98.2  & 93.5 \\
\bottomrule
\end{tabular}
}

\caption{\textbf{Ablation on Training Strategy.} We investigate the different training strategies on pose estimation accuracy.}
    \vspace{-5mm}
\label{tab:abla_train_stra}
\end{table}

%% file: sec/5_conclusion.tex
\section{Conclusion}

We introduced \ours, a feedforward SfM method that can scale up to thousands of input images. It is achieved by augmenting the scene regression Transformer with localization capabilities. By computing a neural scene representation from a subset of anchor images, we fine-tune the Transformer for localization conditioned on the neural scene representation. In this way, the fine-tuned Transformer can quickly reconstruct camera poses and scene points for all the input images. 
Experiments on various benchmarks show state-of-the-art results in both pose estimation and novel view synthesis, surpassing traditional and learning-based baselines in accuracy and efficiency.

\noindent
\textbf{Future  Work \& Limitation.} While our model demonstrates strong performance, two key limitations remain. 
First, global pose estimation in a pre-fixed reference coordinate system might lead to a performance drop on some sequences. A better view selection criterion could improve results.
Second, uniform anchor image sampling risks missing large or diverse scene regions. We could explore coverage-aware selection that maximizes visibility. 

%% file: sec/X_suppl.tex
\clearpage
\setcounter{page}{1}
\maketitlesupplementary

In this supplementary material, we provide additional implementation details and experimental setups in Sec.\ref{sec:more_imple}. We also present further experiments and discussions in Sec.\ref{sec:more_results}. More visualization result will be posted in Sec.~\ref{sec:more_vis}.

\section{More Implementation Details}
\label{sec:more_imple}

\noindent
\textbf{Training Details.} As described, we follow the training set of VGGT~\citep{wang2025vggt}, which we use a cosine learning rate scheduler with a maximum learning rate of $2\times10^{-4}$ and a warm-up of $2$K iterations. The input images are resized to a maximum of $518$ pixels while preserving the aspect ratios between $[0.33,1.0]$.  Data augmentation includes random color jittering, Gaussian blur, and grayscale conversion.   Training is performed with bfloat16 precision, gradient checkpointing, and a mixed anchor–query frame strategy.

\noindent
\textbf{Implementation of $\text{attn}^{\text{query}}$.} We apply an attention mask to realize a cross-attention–like operation between tokens from the query image $\mathcal{I}^{q}$ and the scene representation $\mathcal{R}_j=\Theta(t_j^{'\mathcal{I}})$ from layer $j$. Specifically, during training, the mask enforces two types of interaction:
\begin{itemize}[label=\textbullet]
    \item Anchor interaction: tokens from anchor frames are allowed to attend to each other, enabling mutual information exchange across different anchor views
    \item Query restriction: tokens from query frames cannot attend to tokens from other query frames; they can only attend to tokens within the same frame and to the scene representation $\mathcal{R}_j$.
\end{itemize}
This design ensures that query tokens extract information primarily from the global scene representation and their own local context, while anchor tokens remain fully connected to maximize cross-view aggregation. The same attention mask is also applied to the attention layer in the pose head.We also develop an alternative version that employs two distinct blocks for the anchor and query, respectively. The query block is initialized from the anchor block and fine-tuned during training. We observe that both versions perform similarly.

\noindent
\textbf{Inference Details.} As stated in Sec.\ref{sec:overview}, we first extract the scene representation $\mathcal{R}$ from the anchor frames and then process query images sequentially. Specifically, we employ a KV-cache\citep{radford2019language} to store $\mathcal{R}$ as the keys and values in each global attention layer, which effectively accelerates computation and reduces memory usage. For each subsequent query image, its tokens serve as the queries in the attention mechanism, while the keys/values are formed by concatenating the query tokens with the cached scene tokens. Through the attention operation, the query image tokens are updated by aggregating information from the global scene representation. After passing through all attention layers, we obtain tokens enriched with localization information. These tokens are then fed into the camera head and depth head to predict the corresponding camera parameters, depth and scene coordinate maps, yielding the reconstructed scene from the query viewpoint.

\noindent
\textbf{Post Refinement Details} We adopt Nerfacto~\cite{nerfstudio} within NeRF Studio, applying its camera pose optimizer for post refinement. For scenes with $\leq 2,000$ images, models undergo 10,000 training iterations with a regularization weight $\lambda=0.001$ on both camera translation and rotation. The optimizer uses an initial learning rate of $10^{-3}$, which decays to $10^{-4}$ after $1,000$ iterations via cosine annealing. All other parameters follow the defaults of NeRF Studio. For scenes with $\geq 2,000$ images, we perform two separate optimizations with $10000$ and $30000$ iterations, respectively. The second round uses the poses from the end of the first round as initialization, and the camera optimizer's learning rate begins at $0.0005$ and reduces to $0.00001$. Other parameters remain constant. Optimizations typically take $2.5$ minutes for every 10k iterations, regardless of the number of images.

\noindent
\textbf{More experimental setup.}
\begin{itemize}[label=\textbullet]
    \item We denote DROID-SLAM* as the variant that first calibrates intrinsics using GeoCalib~\citep{veicht2024geocalib} on the first image of each sequence and then uses the calibrated parameters in DROID-SLAM. 
    \item All experiments are conducted on an NVIDIA RTX 4090 GPU; To align with V100-based results, runtimes are scaled by a factor of 1.5, reflecting the measured FP16 inference speed gap. For anchor frame selection, we use 50 frames per scene in 7 scenes, with PSNR reported in the combined training and test sets, and the relocalization accuracy evaluated on the test set after ACE0 \cite{brachmann2024acezero}. For mip-NeRF 360, 50 anchors are selected per sequence. For Tanks and Temples, we select an average of 100 keyframes per sequence to relocalize all remaining images and video frames. For TUM RGB-D, we use 50–100 anchors depending on sequence length: floor, plant, and teddy use 100 frames, and others use 50. 
    \item In cases where the first frame contains limited semantic information, it is replaced with a semantically richer frame as the first anchor.
    \item For visualization in Fig.~\ref{fig:visualization}, we remove points in the lowest 50\% confidence on the depth confidence map, corresponding to sky, glass, and other ambiguous surfaces, and apply moderate point cloud downsampling to enhance visual clarity.
    \item We cite the results in  Tabs.~\ref{tab:sevenscenes_reloc_only}, ~\ref{tab:t2_image_set_psnr}, ~\ref{tab:mip_psnr_only}, ~\ref{tab:supp_t2_pose}, ~\ref{tab:7scenes_psnr} from ACE0~\cite{brachmann2024acezero}. Details on default parameters and configurations for baselines such as COLMAP (default), COLMAP (Sparse + Reloc + BA) and Nope-NeRF are available in the supplementary material of ACE0~\cite{brachmann2024acezero}.
\end{itemize}

\section{Additional Results}
\label{sec:more_results}

\subsection{Pose Estimation}
\input{tables/table_tum_full}
\noindent
\textbf{TUM RGBD.} We report the root mean square error (RMSE) of the absolute trajectory error (ATE), comparing our method with a broader set of state-of-the-art approaches~\cite{campos2021orb, teed2018deepv2d, czarnowski2020deepfactors, lipson2024deep, zhang2023go, teed2021droid, murai2025mast3r} under calibrated settings, as summarized in Tab.~\ref{tab:tum_ate_full}. Our method achieves accuracy on par with the most advanced SLAM systems while remaining robust across diverse sequences without requiring camera calibration. Compared with geometry-based pipelines, such as ORB-SLAM3, our approach exhibits stronger robustness and achieves comparable or superior accuracy to learning-based baselines.
The main weakness appears in the floor sequence, where images contain limited visual cues dominated by textureless floor regions. In this case, reference view selection becomes critical: large viewpoint gaps between the query and reference views significantly degrade localization. We further visualize these effects in the trajectory results (Sec.~\ref{sec:more_vis}).

\subsection{Novel View Synthesis}

\input{tables/table_tnt_psnr}

\noindent
\textbf{Tanks \& Temples.} To further evaluate our scalability for large-scale reconstruction, we apply our method to the Tanks \& Temples~\citep{knapitsch2017tanks} video sequences. For each sequence, we uniformly sample 100 images as anchors and perform localization on all frames.

For clarity, Table~\ref{tab:supp_t2_pose} reports the results on both the image set and the full video sequences, with the latter shown on the right. As a reference, we include \textit{Sparse COLMAP + Reloc + BA} (CMP (SRB)), which initializes from a sparse COLMAP reconstruction using 150–500 images, registers the remaining frames and performs global bundle adjustment.
Our approach consistently outperforms RealityCapture, DROID-SLAM~\citep{teed2021droid}, and ACE0~\citep{brachmann2024acezero} across all splits, with only ACE0 initialized from sparse COLMAP poses (CMP + ACE0) achieving comparable performance. In particular, on the most challenging `advanced' split, our method achieves the highest PSNR of all methods.
Despite each sequence containing more than 10k images on average, our feedforward approach maintains competitive efficiency - only slightly slower than SLAM-based pipelines - while delivering strong reconstruction quality.

\input{tables/table_7scenes_psnr_multi_recon}
\noindent
\textbf{7 Scenes.}
We quote the table from ACE0~\cite{brachmann2024acezero} and report our results in Tab.~\ref{tab:7scenes_psnr}. For each 7-Scenes sequence, we uniformly sample $50$ frames from the train/test splits and estimate poses for all images in the scene. We compare against COLMAP since bundle-adjusted COLMAP poses provide a more accurate reference. Our method attains PSNR comparable to the COLMAP reference and exceeds ACE0. In terms of runtime, even under “fast” settings COLMAP still requires around 13 h per scene; DROID-SLAM returns results quickly but performs poorly on 7-Scenes; ACE0 takes 1 h. In contrast, our approach finishes in 25 min without any pose initialization, achieves higher PSNR than ACE0, and matches the PSNR of ACE0 when initialized from KinectFusion (KF-Init., 7 min). 

To further compare against learning-based approaches under constrained memory budgets,  we follows~\cite{brachmann2024acezero} to downsample each sequence to 200 frames. Sequential-dependent scene regression models (e.g., Cut3R~\citep{wang2025continuous}, SLAM3R~\citep{liu2025slam3r}) require dense temporal input, and VGGT~\cite{wang2025vggt} still exceeds memory limits on 200 images. We therefore compare to BARF~\cite{lin2021barf} and NoPe-NeRF~\cite{bian2022nopenerf}: Both BARF~\cite{lin2021barf} and Nope-NeRF~\cite{bian2022nopenerf} fail to recover the scene after a long fitting time. Using our localization of all frames, we consistently obtain the highest PSNR in this 200-frame setting. while remaining faster than these baselines.

\begin{figure*}
    \centering
    \includegraphics[width=1.0\linewidth]{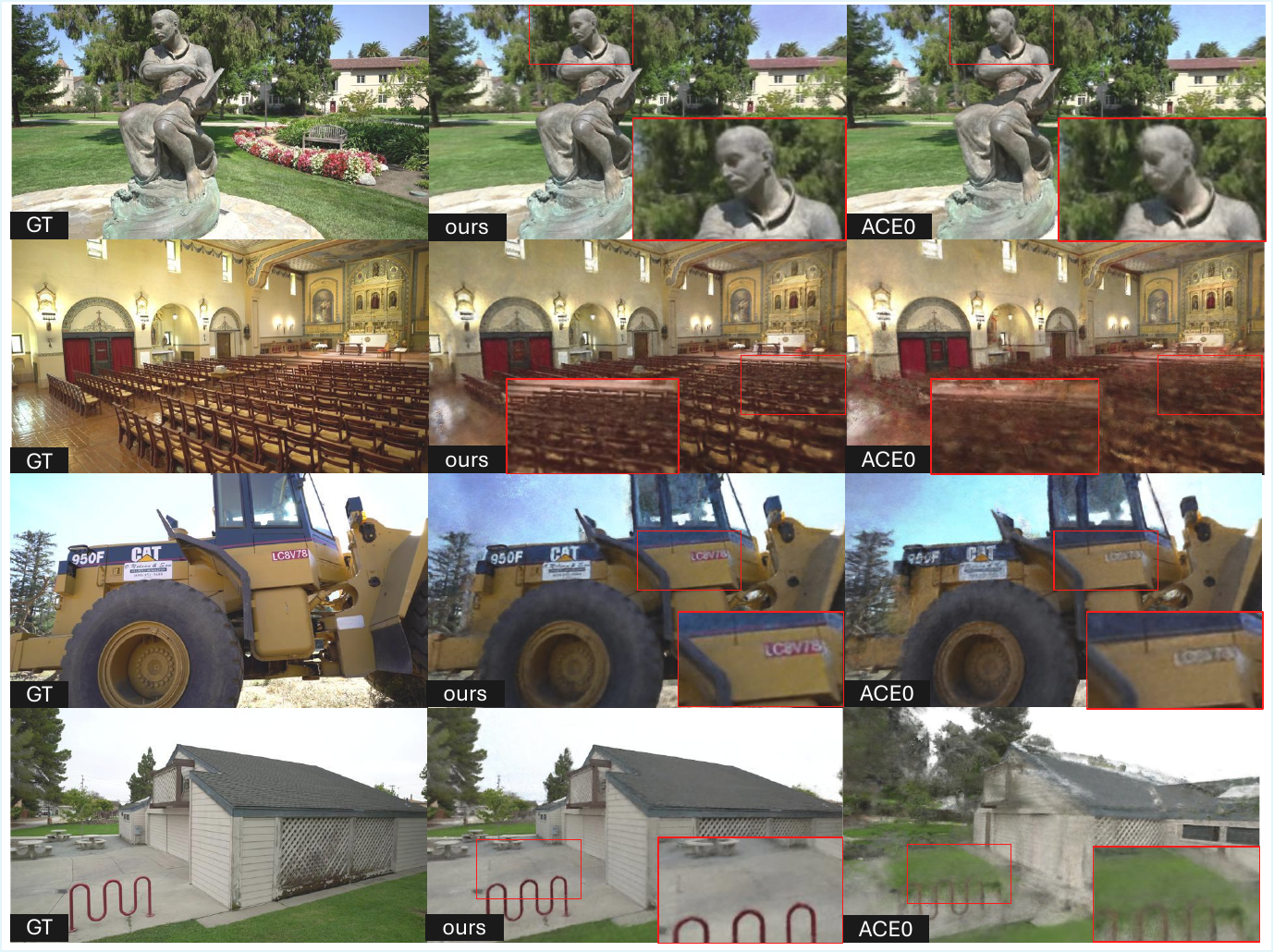}
    \caption{\textbf{Visualization on Tank \& Temple \textit{training} split.}}
    \label{fig:t2_training_psnr_vis}
\end{figure*}

\begin{figure*}
    \centering
    \includegraphics[width=1.0\linewidth]{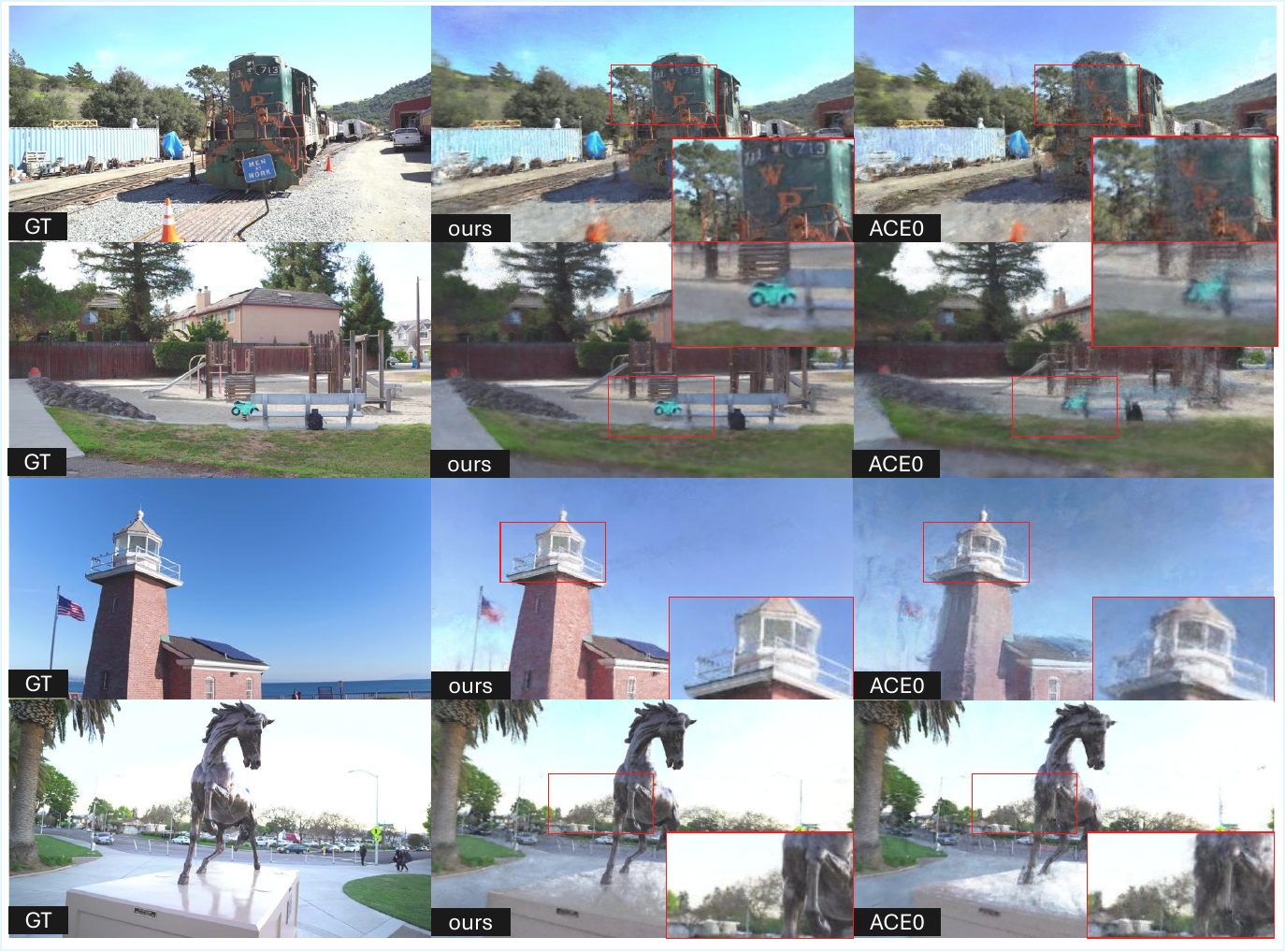}
    \caption{\textbf{Visualization on Tank \& Temple \textit{intermediate} split.}}
    \label{fig:t2_intermediate_psnr_vis}
\end{figure*}

\begin{figure*}
    \centering
    \includegraphics[width=1.0\linewidth]{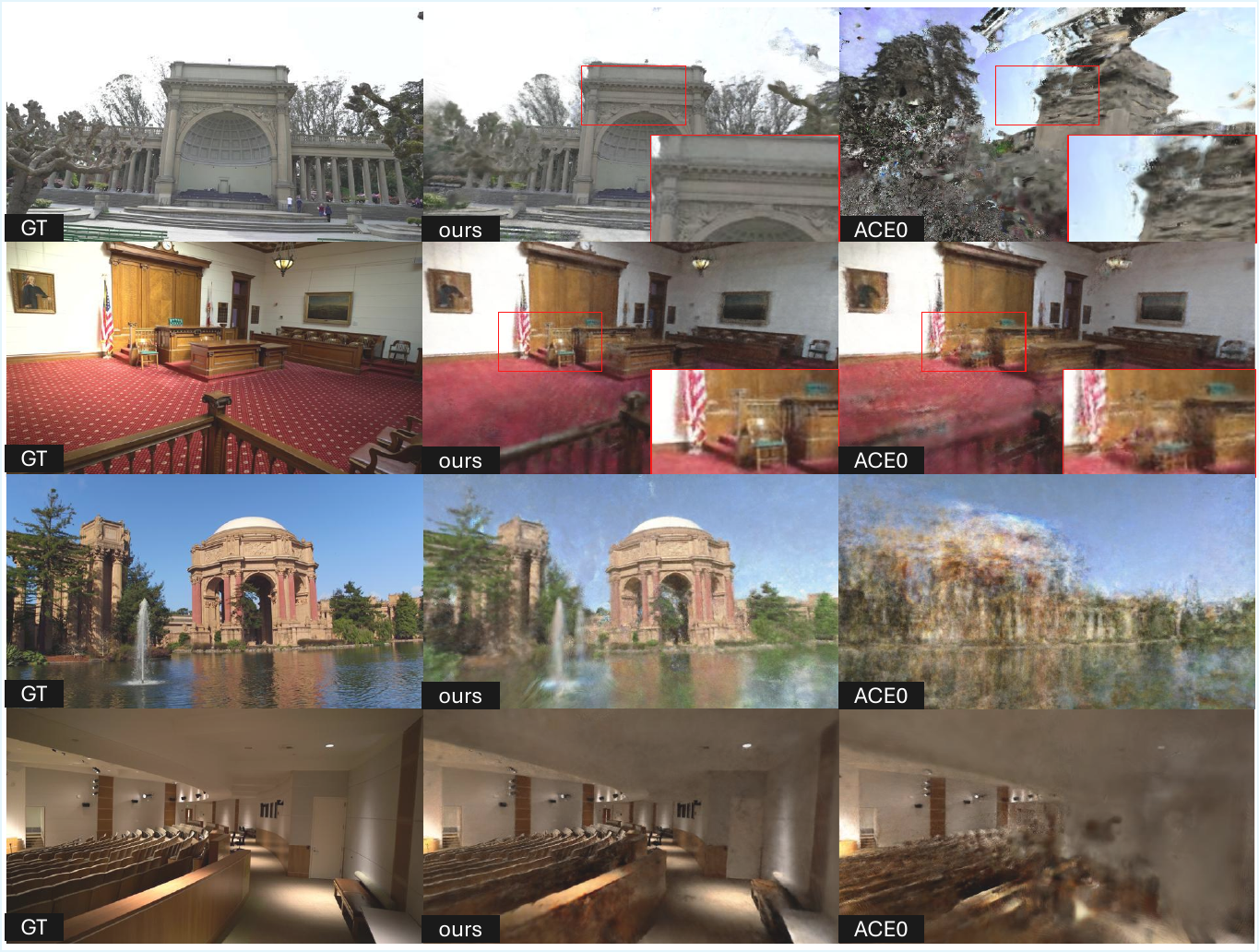}
    \caption{\textbf{Visualization on Tank \& Temple \textit{advanced} split.}}
    \label{fig:t2_advanced_psnr_vis}
\end{figure*}

\begin{figure*}
    \centering
    \includegraphics[width=1.0\linewidth]{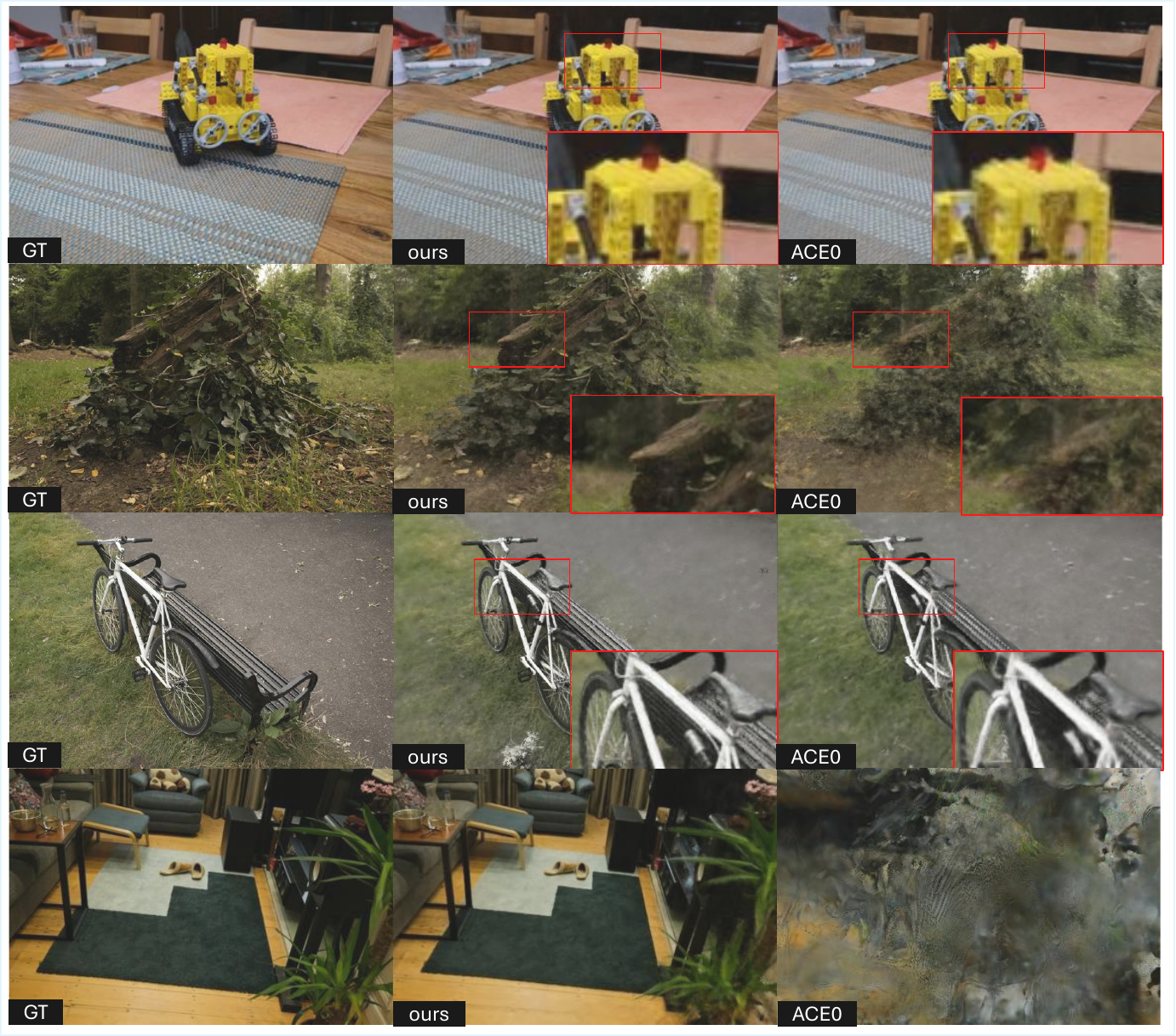}
    \caption{\textbf{Visualization on Mip-NeRF 360 dataset.}}
    \label{fig:t2_mip360_psnr_vis}
\end{figure*}

\begin{figure*}
    \centering
    \includegraphics[width=1.0\linewidth]{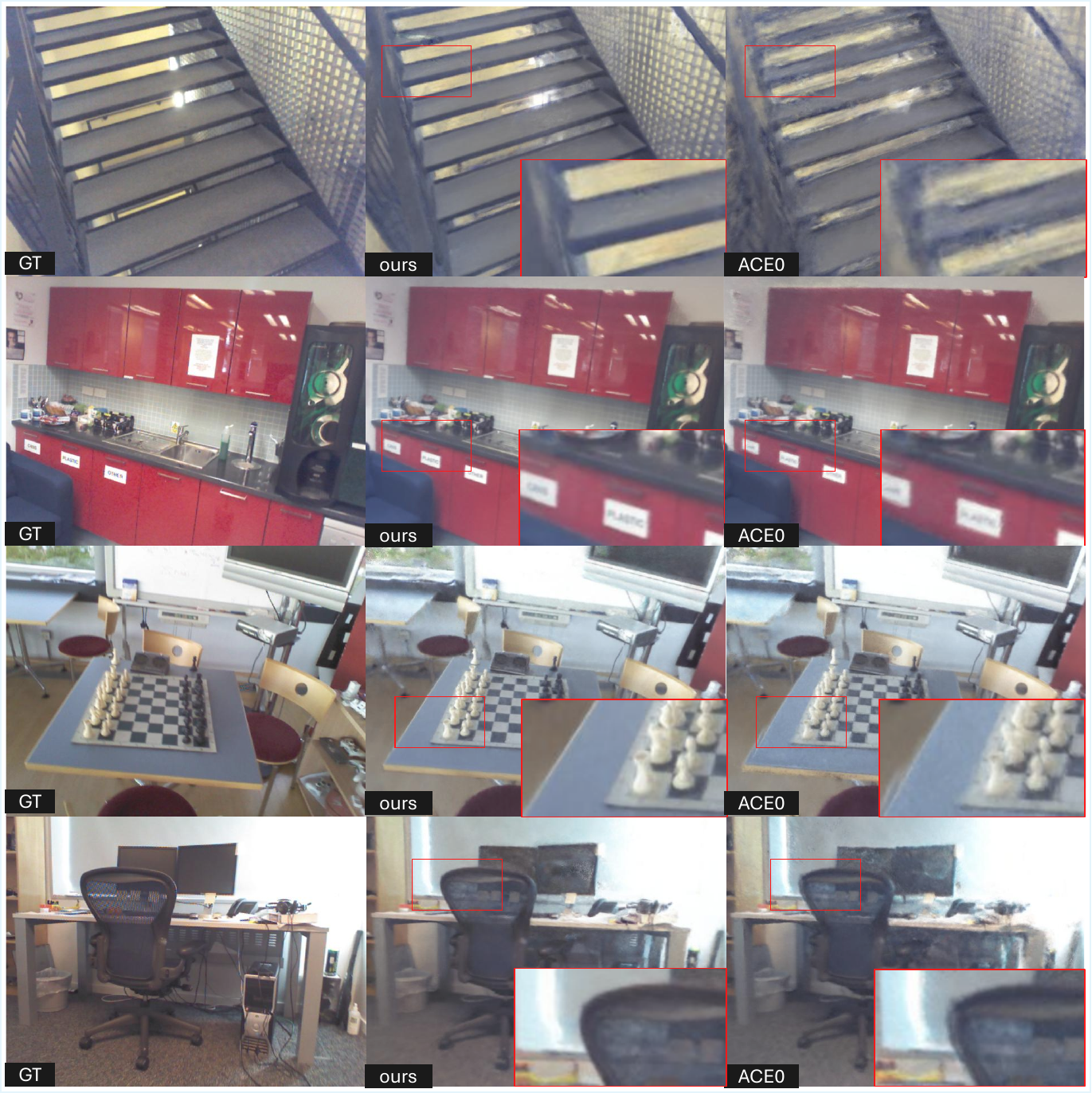}
    \caption{\textbf{Visualization on  7-Scenes dataset.}}
    \label{fig:t2_7scenes_psnr_vis}
\end{figure*}

\begin{figure*}[htbp]
    \centering
    \includegraphics[width=1.0\linewidth]{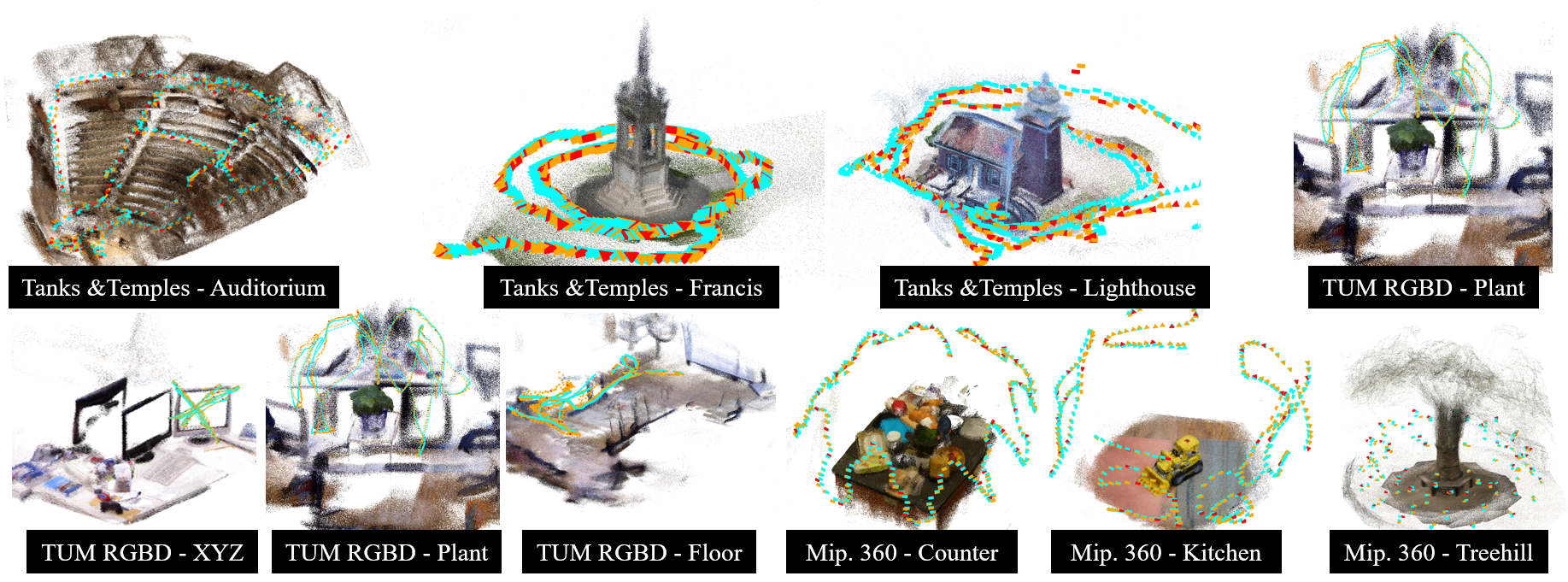}
    \caption{\textbf{Regressed Camera Poses and Point Clouds.} We visualize the camera poses and point clouds predicted by \ours across various datasets. COLMAP or ground-truth camera poses are shown as blue frustums, while regressed camera poses are shown in yellow, with red indicating anchor images.}
    \label{fig:visualization_supp}
    \vspace{-3mm}
\end{figure*}

\section{Visualization}
\label{sec:more_vis}
As shown in Fig.~\ref{fig:t2_training_psnr_vis}, ~\ref{fig:t2_intermediate_psnr_vis} and ~\ref{fig:t2_advanced_psnr_vis}, we show the render images of test view in the three different splits of Tank \& Temple dataset. We illustrate the test view of \textit{7-Scenes} and \textit{Mip-NeRF 360} dataset in Fig.~\ref{fig:t2_mip360_psnr_vis} and ~\ref{fig:t2_7scenes_psnr_vis}, respectively. We aslo supplement in Fig.~\ref{fig:visualization_supp} our regressed camera poses and point clouds.

%% file: tables/table_tum_full.tex
\begin{table*}[h]
    \vspace{-0.5em}
    \centering
    
    \scriptsize
    \begin{tabular}{l|lcccccccccc} 
    \toprule
    \multirow{2}{*}{} & \multirow{2}{*}{Method} & \multicolumn{9}{c}{Sequence} &  \multirow{2}{*}{Avg}  \\ \cmidrule(lr){3-11}
    & &\texttt{360} &\texttt{desk} &\texttt{desk2} &\texttt{floor} &\texttt{plant} &\texttt{room } &\texttt{rpy} &\texttt{teddy} &\texttt{xyz} & \\
    \midrule
    
    \parbox[t]{1mm}{\multirow{8}{*}{\rotatebox[origin=c]{90}{Calib.}}} 
    &  ORB-SLAM3~\cite{campos2021orb} & $  \times$ &  {0.017} &  0.210 &$  \times$ &  0.034 &$   \times$ &  $\times$ &   $\times$ &  \textbf{0.009} &   N/A \\
    &  DeepV2D~\cite{teed2018deepv2d} &  0.243 &  0.166 &  0.379 &  1.653 &  0.203 &  0.246 &  0.105 &  0.316 &  0.064 &  0.375 \\
    &  DeepFactors~\cite{czarnowski2020deepfactors} &  0.159 &  0.170 &  0.253 &  0.169 &  0.305 &  0.364 &  0.043 &  0.601 &  0.035 &  0.233 \\
    &  DPV-SLAM~\cite{lipson2024deep} &  0.112 &  0.018 &  0.029 &  0.057 &  0.021 &  0.330 &  0.030 &  0.084 &  {0.010} &  0.076 \\
    &  DPV-SLAM++~\cite{lipson2024deep} &  0.132 &  0.018 &  0.029 &  0.050 &  0.022 &  0.096 &  0.032 &  0.098 &  {0.010} &  0.054 \\
    &  GO-SLAM~\cite{zhang2023go} &  0.089 &  \textbf{0.016} &  {0.028} &  {0.025} &  0.026 &  {0.052} &  \textbf{0.019} &  0.048 &  {0.010} &  {0.035} \\
    &  DROID-SLAM~\cite{teed2021droid} &  0.111 &  0.018 &  0.042 &  \textbf{0.021} &  \textbf{0.016} &  \textbf{0.049} &  {0.026} &  0.048 &  0.012 &  0.038 \\
    &  \mr-SLAM~\cite{murai2025mast3r} &  \textbf{0.049} &  \textbf{0.016} &  \textbf{0.024} &  {0.025} &  {0.020} &  0.061 &  0.027 &  \textbf{0.041} &  \textbf{0.009} &  \textbf{0.030} \\
    \midrule

    \parbox[t]{1mm}{\multirow{4}{*}{\rotatebox[origin=c]{90}{Uncalib.}}} 
    &DROID-SLAM*~\cite{teed2021droid} &0.202 &  \thirdc 0.032 &0.091 & \secondc 0.064 &0.045 &0.918 &0.056 &0.045 & \firstc 0.012 &0.158 \\
    &\mr-SLAM*~\cite{murai2025mast3r} &\firstc {0.070} & 0.035 & \thirdc 0.055 & \firstc 0.056 &0.035 & 0.118 & \thirdc0.041 &0.114 & 0.020 & \thirdc 0.060 \\
    &VGGT-SLAM~(\Simthree)~\citep{maggio2025vggt} & \thirdc 0.123 & 0.040 & \thirdc 0.055 & 0.254 & \firstc 0.022 & \firstc 0.088 &\thirdc 0.041 & \firstc 0.032 &\thirdc 0.016 &  0.074 \\
    &VGGT-SLAM~(\SLfour)~\citep{maggio2025vggt} & \secondc 0.071 & \secondc 0.025 & \firstc 0.040 & 0.141 & \secondc 0.023 & \secondc 0.102 & \secondc 0.030 & \secondc 0.034 & \secondc 0.014 & \secondc 0.053 \\
    &\ours(Offline)  & \firstc0.070 & \firstc \firstc0.024 & \secondc 0.042 & \thirdc 0.107 & \thirdc 0.031 & \thirdc 0.113 & \firstc 0.020 &  \thirdc0.037 & \firstc 0.012 & \firstc 0.051 \\
    \bottomrule
    \end{tabular}
    \caption{\textbf{Root mean square error~(RMSE) of absolute trajectory error~(ATE) on TUM RGB-D~\citep{sturm2012benchmark} (unit: m)}.
    Gray rows denote results obtained with calibrated camera intrinsics, while entries marked with * indicate evaluation in the uncalibrated setting.  We color result in: \TCBTableBest {Best}, \TCBTableSecond{Second}, and \TCBTableThird{Third}. Note that our method is actually a offline SfM method. }\label{tab:tum_ate_full}
\end{table*}

%% file: tables/table_tnt_psnr.tex
\begin{table*}[t]%
    \centering%
    \setlength{\tabcolsep}{1pt}%
    \footnotesize%
    \adjustbox{valign=t,width=\linewidth}{%
    \begin{tabular}{clcccccccccccccccccccccccccccccccccccc}
         &&&&&&&&&&&&&&&&&&&&&&&&&&&&&&&&&&&&& \\
        \toprule
        \hline
         ~ &%
         ~ &%
        \hspace{-6pt}\parbox[t]{0mm}{\multirow{3}{*}{\rotatebox[origin=c]{90}{Frames}}} &%
        \multicolumn{3}{c|}{~} &%
         & ~ & &%
        \multicolumn{3}{c}{DROID-} &%
        \multicolumn{3}{c}{~} &
        \multicolumn{3}{c||}{~} &%
        \hspace{-6pt}\parbox[t]{0mm}{\multirow{3}{*}{\rotatebox[origin=c]{90}{Frames}}} &%
         & ~ & &%
         \multicolumn{3}{c|}{~} &%
         \multicolumn{3}{c}{~} &
        \multicolumn{3}{c}{DROID-} &%
         & ~ & &%
        \multicolumn{3}{c}{~}\\
         ~ &%
         ~ &%
         ~ &%
        \multicolumn{3}{c|}{CMP} &%
        \multicolumn{3}{c}{Reality} &%
        \multicolumn{3}{c}{SLAM$^\dagger$} &%
        \multicolumn{3}{c}{ACE0} &
        \multicolumn{3}{c||}{Ours} &%
         ~ &%
        \multicolumn{3}{c}{CMP} &%
        \multicolumn{3}{c|}{CMP+} & 
        \multicolumn{3}{c}{Reality} &%
        \multicolumn{3}{c}{SLAM$^\dagger$} &%
        \multicolumn{3}{c}{ACE0}  &
        \multicolumn{3}{c}{~~Ours~~}\\
         ~ &%
         ~ &%
         ~ &%
        \multicolumn{3}{c|}{(D)}  &%
        \multicolumn{3}{c}{Capture} &%
        \multicolumn{3}{c}{\cite{teed2021droid}} &%
        \multicolumn{3}{c}{~~~~} &
        \multicolumn{3}{c||}{~~~~} &%
         ~ &%
        \multicolumn{3}{c}{(SRB)}  &%
        \multicolumn{3}{c|}{~~ACE0 ~~} &
        \multicolumn{3}{c}{Capture} &%
        \multicolumn{3}{c}{\cite{teed2021droid}} &%
        \multicolumn{3}{c}{~~~~} &%
        \multicolumn{3}{c}{~~~~}\\
        \midrule
        \parbox[t]{3mm}{
        \multirow{8}{*}{\rotatebox[origin=c]{90}{Training}}} 
        & Barn & 410 & 
        \multicolumn{3}{c|}{\begin{tabular}{ccc} & ~{24.0}~ & \end{tabular}} &  
         \multicolumn{3}{c}{\begin{tabular}{ccc} & ~{\secondc21.2}~ & \end{tabular}} &  
         \multicolumn{3}{c}{\begin{tabular}{ccc} & ~{\thirdc19.0}~ & \end{tabular}} &  
         \multicolumn{3}{c}{\begin{tabular}{ccc} & ~{16.5}~ & \end{tabular}} &
         \multicolumn{3}{c||}{\begin{tabular}{ccc} &~{\firstc23.5}~ & \end{tabular}}&
         19.3k &         
         \multicolumn{3}{c}{\begin{tabular}{ccc} & ~{26.3}~ & \end{tabular}} &  
         \multicolumn{3}{c|}{\begin{tabular}{ccc} & ~{\firstc25.1}~ & \end{tabular}} &  
         \multicolumn{3}{c}{\begin{tabular}{ccc} & ~{16.9}~ & \end{tabular}} &  
         \multicolumn{3}{c}{\begin{tabular}{ccc} & ~{13.5}~ & \end{tabular}} &  
         \multicolumn{3}{c}{\begin{tabular}{ccc} & ~{\thirdc17.7}~ & \end{tabular}} &  
         \multicolumn{3}{c}{\begin{tabular}{ccc} &~{\firstc25.1}~ & \end{tabular}} \\

        &&&&&&&&&&&&&&&&&&&&&&&&&&&&&&&&&&&&&\\[-10pt]
         & Catpr. & 383 &  
         \multicolumn{3}{c|}{\begin{tabular}{ccc} & ~{17.1}~ & \end{tabular}} &  
         \multicolumn{3}{c}{\begin{tabular}{ccc} & ~{15.9}~ & \end{tabular}} &  
         \multicolumn{3}{c}{\begin{tabular}{ccc} & ~{\thirdc16.6}~ & \end{tabular}} &  
         \multicolumn{3}{c}{\begin{tabular}{ccc} & ~{\firstc16.9}~ & \end{tabular}} &
         \multicolumn{3}{c||}{\begin{tabular}{ccc}&~{\secondc16.8}~ & \end{tabular}} & 
         11.4k &  
         \multicolumn{3}{c}{\begin{tabular}{ccc} & ~{18.7}~ & \end{tabular}} &   
         \multicolumn{3}{c|}{\begin{tabular}{ccc} & ~{\secondc18.8}~ & \end{tabular}} & 
         \multicolumn{3}{c}{\begin{tabular}{ccc} & ~{17.9}~ & \end{tabular}} &  
         \multicolumn{3}{c}{\begin{tabular}{ccc} & ~{\firstc18.9}~ & \end{tabular}} &  
         \multicolumn{3}{c}{\begin{tabular}{ccc} & ~{\thirdc18.6}~ & \end{tabular}}  & 
         \multicolumn{3}{c}{\begin{tabular}{ccc} &~{17.5}~ & \end{tabular}} \\
        &&&&&&&&&&&&&&&&&&&&&&&&&&&&&&&&&&&&&\\[-10pt]
         & Church & 507 & 
        \multicolumn{3}{c|}{\begin{tabular}{ccc} & ~{18.3}~ & \end{tabular}} &  
         \multicolumn{3}{c}{\begin{tabular}{ccc} & ~{\firstc17.6}~ & \end{tabular}} &  
         \multicolumn{3}{c}{\begin{tabular}{ccc} & ~{14.3}~ & \end{tabular}} &  
         \multicolumn{3}{c}{\begin{tabular}{ccc} & ~{\secondc17.2}~ & \end{tabular}} &
         \multicolumn{3}{c||}{\begin{tabular}{ccc} &~{\thirdc17.0}~ & \end{tabular}}&
         19.3k &         
         \multicolumn{3}{c}{\begin{tabular}{ccc} & ~{18.5}~ & \end{tabular}} &  
         \multicolumn{3}{c|}{\begin{tabular}{ccc} & ~{\firstc17.3}~ & \end{tabular}} &  
         \multicolumn{3}{c}{\begin{tabular}{ccc} & ~-~ & \end{tabular}} &  
         \multicolumn{3}{c}{\begin{tabular}{ccc} & ~{11.5}~ & \end{tabular}} &  
         \multicolumn{3}{c}{\begin{tabular}{ccc} & ~{\secondc16.5}~ & \end{tabular}} &  
         \multicolumn{3}{c}{\begin{tabular}{ccc} &~{\thirdc15.8}~ & \end{tabular}} \\
        &&&&&&&&&&&&&&&&&&&&&&&&&&&&&&&&&&&&&\\[-10pt]
         & Ignatius & 264 & 
         \multicolumn{3}{c|}{\begin{tabular}{ccc} & ~{20.1}~ & \end{tabular}} &  
         \multicolumn{3}{c}{\begin{tabular}{ccc} & ~{17.7}~ & \end{tabular}} &  
         \multicolumn{3}{c}{\begin{tabular}{ccc} & ~{\thirdc17.8}~ & \end{tabular}} &  
         \multicolumn{3}{c}{\begin{tabular}{ccc} & ~{\firstc19.8}~ & \end{tabular}} &
         \multicolumn{3}{c||}{\begin{tabular}{ccc} &~{\secondc19.5}~ & \end{tabular}}& 
         7.8k & 
         \multicolumn{3}{c}{\begin{tabular}{ccc} & ~{20.9}~ & \end{tabular}} &  
         \multicolumn{3}{c|}{\begin{tabular}{ccc} & ~{\firstc20.7}~ & \end{tabular}}&
         \multicolumn{3}{c}{\begin{tabular}{ccc} & ~{\thirdc18.6}~ & \end{tabular}} &  
         \multicolumn{3}{c}{\begin{tabular}{ccc} & ~{\secondc19.1}~ & \end{tabular}} &  
         \multicolumn{3}{c}{\begin{tabular}{ccc} & ~{\firstc20.7}~ & \end{tabular}} &  
         \multicolumn{3}{c}{\begin{tabular}{ccc} &~{\firstc20.7}~ & \end{tabular}}\\

            &&&&&&&&&&&&&&&&&&&&&&&&&&&&&&&&&&&&&\\[-10pt]
         & MtgRm. & 371 &
          \multicolumn{3}{c|}{\begin{tabular}{ccc} & ~{18.6}~ & \end{tabular}} &
         \multicolumn{3}{c}{\begin{tabular}{ccc} & ~{\secondc18.1}~ & \end{tabular}} &  
         \multicolumn{3}{c}{\begin{tabular}{ccc} & ~{15.6}~ & \end{tabular}} &  
         \multicolumn{3}{c}{\begin{tabular}{ccc} & ~{\thirdc18.0}~ & \end{tabular}} &
         \multicolumn{3}{c||}{\begin{tabular}{ccc} &~{\firstc19.5}~ & \end{tabular}} &
         11.1k &  
         \multicolumn{3}{c}{\begin{tabular}{ccc} & ~{20.8}~ & \end{tabular}} &  
         \multicolumn{3}{c|}{\begin{tabular}{ccc} & ~{\secondc20.3}~ & \end{tabular}} & 
         \multicolumn{3}{c}{\begin{tabular}{ccc} & ~{\thirdc18.2}~ & \end{tabular}} & 
         \multicolumn{3}{c}{\begin{tabular}{ccc} & ~{17.1}~ & \end{tabular}} &  
         \multicolumn{3}{c}{\begin{tabular}{ccc} & ~{16.6}~ & \end{tabular}} &  
         \multicolumn{3}{c}{\begin{tabular}{ccc} &~{\firstc20.4}~ & \end{tabular}} \\

        &&&&&&&&&&&&&&&&&&&&&&&&&&&&&&&&&&&&&\\[-10pt]
         & Truck & 251 &
         \multicolumn{3}{c|}{\begin{tabular}{ccc} & ~{21.1}~ & \end{tabular}} & 
         \multicolumn{3}{c}{\begin{tabular}{ccc} & ~{\thirdc19.0}~ & \end{tabular}} &  
         \multicolumn{3}{c}{\begin{tabular}{ccc} & ~{18.3}~ & \end{tabular}} &  
         \multicolumn{3}{c}{\begin{tabular}{ccc} & ~{\secondc20.1}~ & \end{tabular}} & 
         \multicolumn{3}{c||}{\begin{tabular}{ccc} &~{\firstc20.9}~ & \end{tabular}}&
         7.5k &
         \multicolumn{3}{c}{\begin{tabular}{ccc} & ~{23.4}~ & \end{tabular}} & 
         \multicolumn{3}{c|}{\begin{tabular}{ccc} & ~{\secondc23.1}~ & \end{tabular}} & 
         \multicolumn{3}{c}{\begin{tabular}{ccc} & ~{19.1}~ & \end{tabular}} &  
         \multicolumn{3}{c}{\begin{tabular}{ccc} & ~{20.6}~ & \end{tabular}} &  
         \multicolumn{3}{c}{\begin{tabular}{ccc} & ~{\thirdc23.0}~ & \end{tabular}} &  
         \multicolumn{3}{c}{\begin{tabular}{ccc} &~{\firstc23.5}~ & \end{tabular}}\\

        &&&&&&&&&&&&&&&&&&&&&&&&&&&&&&&&&&&&&\\[-10pt]
        \hhline{~-----------------------------------}
    &&&&&&&&&&&&&&&&&&&&&&&&&&&&&&&&&&&&&\\[-10pt]
         & Average & 364 &
         \multicolumn{3}{c|}{\begin{tabular}{ccc} & ~{19.9}~ & \end{tabular}} &  
         \multicolumn{3}{c}{\begin{tabular}{ccc} & ~{\secondc18.2}~ & \end{tabular}} &  
         \multicolumn{3}{c}{\begin{tabular}{ccc} & ~{16.9}~ & \end{tabular}} &  
         \multicolumn{3}{c}{\begin{tabular}{ccc} & ~{\thirdc18.1}~ & \end{tabular}} & 
         \multicolumn{3}{c||}{\begin{tabular}{ccc} &~{\firstc19.5}~ & \end{tabular}}& 
         14.6k & 
         \multicolumn{3}{c}{\begin{tabular}{ccc} & ~{21.4}~ & \end{tabular}} &  
         \multicolumn{3}{c|}{\begin{tabular}{ccc} & ~{\firstc20.9}~ & \end{tabular}}  & 
         \multicolumn{3}{c}{\begin{tabular}{ccc} & ~{18.2}~ & \end{tabular}} &  
         \multicolumn{3}{c}{\begin{tabular}{ccc} & ~{16.8}~ & \end{tabular}} &  
         \multicolumn{3}{c}{\begin{tabular}{ccc} & ~{\thirdc18.9}~ & \end{tabular}} &  
         \multicolumn{3}{c}{\begin{tabular}{ccc} &~{\secondc20.5}~ & \end{tabular}}\\

     &&&&&&&&&&&&&&&&&&&&&&&&&&&&&&&&&&&&&\\[-10pt]
        \hline
         & Time &      & 
         \multicolumn{3}{c|}{\begin{tabular}{ccc} & ~1h~ & \end{tabular}} &
         \multicolumn{3}{c}{\begin{tabular}{ccc} & ~3min~ & \end{tabular}} &  
         \multicolumn{3}{c}{\begin{tabular}{ccc} & ~5min~ & \end{tabular}} &  
         \multicolumn{3}{c}{\begin{tabular}{ccc} & ~1.1h~ & \end{tabular}} &  
         \multicolumn{3}{c||}{\begin{tabular}{ccc} & ~3.5min~ & \end{tabular}} &     &          
         \multicolumn{3}{c}{\begin{tabular}{ccc} & ~8h~ & \end{tabular}} &  
         \multicolumn{3}{c|}{\begin{tabular}{ccc} & ~1.8h~ & \end{tabular}} &   
         \multicolumn{3}{c}{\begin{tabular}{ccc} & ~14h~ & \end{tabular}} &  
         \multicolumn{3}{c}{\begin{tabular}{ccc} & ~18min~ & \end{tabular}} &  
         \multicolumn{3}{c}{\begin{tabular}{ccc} & ~2.2h~ & \end{tabular}} &  
         \multicolumn{3}{c}{\begin{tabular}{ccc} & ~58min~ & \end{tabular}} \\

        \hline
         &&&&&&&&&&&&&&&&&&&&&&&&&&&&&&&&&&&&& \\
        \hline

          &&&&&&&&&&&&&&&&&&&&&&&&&&&&&&&&&&&&&  \\[-10pt]
        \parbox[t]{5mm}{\multirow{8}{*}{\rotatebox[origin=c]{90}{Intermediate}}} 
        & Family & 152 & 
         \multicolumn{3}{c|}{\begin{tabular}{ccc} & ~{19.5}~ & \end{tabular}} & 
        \multicolumn{3}{c}{\begin{tabular}{ccc} & ~{\thirdc18.8}~ & \end{tabular}} & 
        \multicolumn{3}{c}{\begin{tabular}{ccc} & ~{17.6}~ & \end{tabular}} &  
        \multicolumn{3}{c}{\begin{tabular}{ccc} & ~{\secondc19.0}~ & \end{tabular}} &
        \multicolumn{3}{c||}{\begin{tabular}{ccc} &~{\firstc20.6}~ & \end{tabular}}& 
        4.4k & 
        \multicolumn{3}{c}{\begin{tabular}{ccc} & ~{21.3}~ & \end{tabular}} &  
        \multicolumn{3}{c|}{\begin{tabular}{ccc} & ~{\firstc21.3}~ & \end{tabular}} & 
        \multicolumn{3}{c}{\begin{tabular}{ccc} & ~{\secondc19.8}~ & \end{tabular}} &  
        \multicolumn{3}{c}{\begin{tabular}{ccc} & ~{\secondc19.8}~ & \end{tabular}} &  
        \multicolumn{3}{c}{\begin{tabular}{ccc} & ~{\thirdc18.0}~ & \end{tabular}} &  
        \multicolumn{3}{c}{\begin{tabular}{ccc} &~{\firstc21.3}~ & \end{tabular}} \\
        
        &&&&&&&&&&&&&&&&&&&&&&&&&&&&&&&&&&&&&\\[-10pt]
         & Francis & 302 &
         \multicolumn{3}{c|}{\begin{tabular}{ccc} & ~{21.6}~ & \end{tabular}} &    
         \multicolumn{3}{c}{\begin{tabular}{ccc} & ~{\secondc20.7}~ & \end{tabular}} &  
         \multicolumn{3}{c}{\begin{tabular}{ccc} & ~{\secondc20.7}~ & \end{tabular}} &  
         \multicolumn{3}{c}{\begin{tabular}{ccc} & ~{\thirdc20.1}~ & \end{tabular}} &
         \multicolumn{3}{c||}{\begin{tabular}{ccc} &~{\firstc21.8}~ & \end{tabular}} &
         7.8k & 
         \multicolumn{3}{c}{\begin{tabular}{ccc} & ~{22.5}~ & \end{tabular}} & 
         \multicolumn{3}{c|}{\begin{tabular}{ccc} & ~{\secondc  22.7}~ & \end{tabular}} &  
         \multicolumn{3}{c}{\begin{tabular}{ccc} & ~{20.4}~ & \end{tabular}} & 
         \multicolumn{3}{c}{\begin{tabular}{ccc} & ~{\thirdc  21.8}~ & \end{tabular}} &  
         \multicolumn{3}{c}{\begin{tabular}{ccc} & ~{21.7}~ & \end{tabular}} & 
         \multicolumn{3}{c}{\begin{tabular}{ccc} &~{\firstc22.8}~ & \end{tabular}} \\

         &&&&&&&&&&&&&&&&&&&&&&&&&&&&&&&&&&&&&\\[-10pt]
         & Horse & 151 & 
         \multicolumn{3}{c|}{\begin{tabular}{ccc} & ~{19.2}~ & \end{tabular}} & 
         \multicolumn{3}{c}{\begin{tabular}{ccc} & ~{\thirdc19.0}~ & \end{tabular}} &  
         \multicolumn{3}{c}{\begin{tabular}{ccc} & ~{16.3}~ & \end{tabular}} &  
         \multicolumn{3}{c}{\begin{tabular}{ccc} & ~{\secondc19.5}~ & \end{tabular}} &
         \multicolumn{3}{c||}{\begin{tabular}{ccc} &~{\firstc20.1}~ & \end{tabular}} & 
         6.0k & 
         \multicolumn{3}{c}{\begin{tabular}{ccc} & ~{22.6}~ & \end{tabular}} &  
         \multicolumn{3}{c|}{\begin{tabular}{ccc} & ~{\firstc 22.3}~ & \end{tabular}} & 
         \multicolumn{3}{c}{\begin{tabular}{ccc} & ~{20.7}~ & \end{tabular}} &  
         \multicolumn{3}{c}{\begin{tabular}{ccc} & ~{19.2}~ & \end{tabular}} &  
         \multicolumn{3}{c}{\begin{tabular}{ccc} & ~{\thirdc21.7}~ & \end{tabular}} &  
         \multicolumn{3}{c}{\begin{tabular}{ccc} &~{\secondc21.9}~ & \end{tabular}}\\

         &&&&&&&&&&&&&&&&&&&&&&&&&&&&&&&&&&&&&\\[-10pt]
         & LightH. & 309 & 
         \multicolumn{3}{c|}{\begin{tabular}{ccc} & ~{16.6}~ & \end{tabular}} &   
         \multicolumn{3}{c}{\begin{tabular}{ccc} & ~{\thirdc16.5}~ & \end{tabular}} & 
         \multicolumn{3}{c}{\begin{tabular}{ccc} & ~{13.6}~ & \end{tabular}} &  
         \multicolumn{3}{c}{\begin{tabular}{ccc} & ~{\secondc17.5}~ & \end{tabular}} &
         \multicolumn{3}{c||}{\begin{tabular}{ccc} &~{\firstc18.2}~ & \end{tabular}} &
         8.3k & 
         \multicolumn{3}{c}{\begin{tabular}{ccc} & ~{19.5}~ & \end{tabular}} &  
         \multicolumn{3}{c|}{\begin{tabular}{ccc} & ~{\firstc 20.5}~ & \end{tabular}}  & 
         \multicolumn{3}{c}{\begin{tabular}{ccc} & ~{16.6}~ & \end{tabular}} & 
         \multicolumn{3}{c}{\begin{tabular}{ccc} & ~{\thirdc 18.9}~ & \end{tabular}} &  
         \multicolumn{3}{c}{\begin{tabular}{ccc} & ~{18.6}~ & \end{tabular}} &  
         \multicolumn{3}{c}{\begin{tabular}{ccc} &~{\secondc19.7}~ & \end{tabular}} \\

         &&&&&&&&&&&&&&&&&&&&&&&&&&&&&&&&&&&&&\\[-10pt]
         & PlayGd. & 307 & 
         \multicolumn{3}{c|}{\begin{tabular}{ccc} & ~{19.1}~ & \end{tabular}} &  
         \multicolumn{3}{c}{\begin{tabular}{ccc} & ~{\secondc19.2}~ & \end{tabular}} &  
         \multicolumn{3}{c}{\begin{tabular}{ccc} & ~{11.4}~ & \end{tabular}} &  
         \multicolumn{3}{c}{\begin{tabular}{ccc} & ~{\thirdc18.7}~ & \end{tabular}} & 
         \multicolumn{3}{c||}{\begin{tabular}{ccc} &~{\firstc20.3}~ & \end{tabular}} & 
         7.7k & 
        \multicolumn{3}{c}{\begin{tabular}{ccc} & ~{21.2}~ & \end{tabular}} &  
        \multicolumn{3}{c|}{\begin{tabular}{ccc} & ~{\secondc21.0}~ & \end{tabular}} & 
         \multicolumn{3}{c}{\begin{tabular}{ccc} & ~{16.5}~ & \end{tabular}} &  
         \multicolumn{3}{c}{\begin{tabular}{ccc} & ~{11.3}~ & \end{tabular}} &  
         \multicolumn{3}{c}{\begin{tabular}{ccc} & ~{\thirdc 20.4}~ & \end{tabular}} &  
         \multicolumn{3}{c}{\begin{tabular}{ccc} &~{\firstc21.7}~ & \end{tabular}} \\

 &&&&&&&&&&&&&&&&&&&&&&&&&&&&&&&&&&&&&\\[-10pt]
         & Train & 301 & 
         \multicolumn{3}{c|}{\begin{tabular}{ccc} & ~{16.8}~ & \end{tabular}} & 
         \multicolumn{3}{c}{\begin{tabular}{ccc} & ~{\secondc15.4}~ & \end{tabular}} &  
         \multicolumn{3}{c}{\begin{tabular}{ccc} & ~{\thirdc13.8}~ & \end{tabular}} &  
         \multicolumn{3}{c}{\begin{tabular}{ccc} & ~{\firstc16.2}~ & \end{tabular}} & 
         \multicolumn{3}{c||}{\begin{tabular}{ccc} &~{\firstc16.2}~ & \end{tabular}} &
         12.6k & 
        \multicolumn{3}{c}{\begin{tabular}{ccc} & ~{19.8}~ & \end{tabular}} & 
        \multicolumn{3}{c|}{\begin{tabular}{ccc} & ~{\firstc18.5}~ & \end{tabular}} & 
         \multicolumn{3}{c}{\begin{tabular}{ccc} & ~{\thirdc14.4}~ & \end{tabular}} &  
         \multicolumn{3}{c}{\begin{tabular}{ccc} & ~{\secondc15.6}~ & \end{tabular}} &  
         \multicolumn{3}{c}{\begin{tabular}{ccc} & ~{\firstc18.5}~ & \end{tabular}} & 
         \multicolumn{3}{c}{\begin{tabular}{ccc} &~{\firstc18.5}~ & \end{tabular}} \\

         & & & & & & & & & & & & & & & & & & & & & & & & & & & & & \\[-10pt]
        \hhline{~-----------------------------------}
         & & & & & & & & & & & & & & & & & & & & & & & & & & & & & \\[-10pt]
         & Average & 254 & 
         \multicolumn{3}{c|}{\begin{tabular}{ccc} & ~{18.8}~ & \end{tabular}} &  
         \multicolumn{3}{c}{\begin{tabular}{ccc} & ~{\thirdc18.3}~ & \end{tabular}} &  
         \multicolumn{3}{c}{\begin{tabular}{ccc} & ~{15.6}~ & \end{tabular}} &  
         \multicolumn{3}{c}{\begin{tabular}{ccc} & ~{\secondc18.5}~ & \end{tabular}} & 
         \multicolumn{3}{c||}{\begin{tabular}{ccc} &~{\firstc19.5}~ & \end{tabular}} & 
         7.8k & 
         \multicolumn{3}{c}{\begin{tabular}{ccc} & ~{21.1}~ & \end{tabular}} & 
         \multicolumn{3}{c|}{\begin{tabular}{ccc} & ~{\firstc21.0}~ & \end{tabular}}  & 
         \multicolumn{3}{c}{\begin{tabular}{ccc} & ~{\thirdc18.1}~ & \end{tabular}} &  
         \multicolumn{3}{c}{\begin{tabular}{ccc} & ~{17.8}~ & \end{tabular}} &  
         \multicolumn{3}{c}{\begin{tabular}{ccc} & ~{\secondc19.8}~ & \end{tabular}} & 
         \multicolumn{3}{c}{\begin{tabular}{ccc} &~{\firstc21.0}~ & \end{tabular}}\\

         & & & & & & & & & & & & & & & & & & & & & & & & & & & & & \\[-10pt]
        \hline
         & Time &      & 
         \multicolumn{3}{c|}{\begin{tabular}{ccc} & ~32min~ & \end{tabular}} &  
         \multicolumn{3}{c}{\begin{tabular}{ccc} & ~2min~ & \end{tabular}} &  
         \multicolumn{3}{c}{\begin{tabular}{ccc} & ~3min~ & \end{tabular}} &  
         \multicolumn{3}{c}{\begin{tabular}{ccc} & ~1.3h~ & \end{tabular}} &   
         \multicolumn{3}{c||}{\begin{tabular}{ccc} & ~3min~ & \end{tabular}} &    & 
         \multicolumn{3}{c}{\begin{tabular}{ccc} & ~5h~ & \end{tabular}} &  
         \multicolumn{3}{c|}{\begin{tabular}{ccc} & ~1h~ & \end{tabular}} &  
         \multicolumn{3}{c}{\begin{tabular}{ccc} & ~11h~ & \end{tabular}} & 
         \multicolumn{3}{c}{\begin{tabular}{ccc} & ~14min~ & \end{tabular}} &  
         \multicolumn{3}{c}{\begin{tabular}{ccc} & ~2.2h~ & \end{tabular}} &  
         \multicolumn{3}{c}{\begin{tabular}{ccc} & ~30min~ & \end{tabular}}\\
        \hline
         & & & & & & & & & & & & & & & & & & & & & & & & & & & & & \\
        \hline
        
         & & & & & & & & & & & & & & & & & & & & & & & & & & & & & \\[-10pt]
    \parbox[t]{5mm}{\multirow{7}{*}{\rotatebox[origin=c]{90}{Advanced}}} & 
    Audtrm. & 302 & 
    \multicolumn{3}{c|}{\begin{tabular}{ccc} & ~{19.6}~ & \end{tabular}} & 
    \multicolumn{3}{c}{\begin{tabular}{ccc} & ~{12.2}~ & \end{tabular}} & 
    \multicolumn{3}{c}{\begin{tabular}{ccc} & ~{\thirdc16.7}~ & \end{tabular}} & 
    \multicolumn{3}{c}{\begin{tabular}{ccc} & ~{\secondc18.7}~ & \end{tabular}} & 
    \multicolumn{3}{c||}{\begin{tabular}{ccc} &~{\firstc20.3}~ & \end{tabular}} & 
    13.6k & 
    \multicolumn{3}{c}{\begin{tabular}{ccc} & ~{21.4}~ & \end{tabular}} &  
    \multicolumn{3}{c|}{\begin{tabular}{ccc} & ~{\thirdc 19.8}~ & \end{tabular}} & 
    \multicolumn{3}{c}{\begin{tabular}{ccc} & ~-~ & \end{tabular}} & 
    \multicolumn{3}{c}{\begin{tabular}{ccc} & ~{16.6}~ & \end{tabular}} &  
    \multicolumn{3}{c}{\begin{tabular}{ccc} & ~{\secondc20.0}~ & \end{tabular}} &  
    \multicolumn{3}{c}{\begin{tabular}{ccc} &~{\firstc21.0}~ & \end{tabular}} \\

         & & & & & & & & & & & & & & & & & & & & & & & & & & & & & \\[-10pt]
         & BallRm. & 324 & 
         \multicolumn{3}{c|}{\begin{tabular}{ccc} & ~{16.3}~ & \end{tabular}} &  
         \multicolumn{3}{c}{\begin{tabular}{ccc} & ~{\firstc 18.3}~ & \end{tabular}} & 
         \multicolumn{3}{c}{\begin{tabular}{ccc} & ~{13.1}~ & \end{tabular}} &  
         \multicolumn{3}{c}{\begin{tabular}{ccc} & ~{\secondc17.9}~ & \end{tabular}} &
         \multicolumn{3}{c||}{\begin{tabular}{ccc} &~{\thirdc 14.8}~ & \end{tabular}} & 
         10.8k & 
        \multicolumn{3}{c}{\begin{tabular}{ccc} & ~{18.0}~ & \end{tabular}} & 
        \multicolumn{3}{c|}{\begin{tabular}{ccc} & ~{\thirdc 15.6}~ & \end{tabular}} & 
         \multicolumn{3}{c}{\begin{tabular}{ccc} & ~-~ & \end{tabular}} &  
         \multicolumn{3}{c}{\begin{tabular}{ccc} & ~{10.4}~ & \end{tabular}} &  
         \multicolumn{3}{c}{\begin{tabular}{ccc} & ~{\firstc 18.9}~ & \end{tabular}} &  
         \multicolumn{3}{c}{\begin{tabular}{ccc} &~{\secondc16.9}~ & \end{tabular}} \\

         & & & & & & & & & & & & & & & & & & & & & & & & & & & & & \\[-10pt]
         & CortRm. & 301 & 
         \multicolumn{3}{c|}{\begin{tabular}{ccc} & ~{18.2}~ & \end{tabular}} &  
         \multicolumn{3}{c}{\begin{tabular}{ccc} & ~{\secondc17.2}~ & \end{tabular}} & 
         \multicolumn{3}{c}{\begin{tabular}{ccc} & ~{12.3}~ & \end{tabular}} & 
         \multicolumn{3}{c}{\begin{tabular}{ccc} & ~{\thirdc17.1}~ & \end{tabular}} & 
         \multicolumn{3}{c||}{\begin{tabular}{ccc} &~{\firstc17.4}~ & \end{tabular}} & 
         12.6k & 
        \multicolumn{3}{c}{\begin{tabular}{ccc} & ~{18.7}~ & \end{tabular}} & 
        \multicolumn{3}{c|}{\begin{tabular}{ccc} & ~{\firstc17.8}~ & \end{tabular}} & 
         \multicolumn{3}{c}{\begin{tabular}{ccc} & ~-~ & \end{tabular}} & 
         \multicolumn{3}{c}{\begin{tabular}{ccc} & ~{10.2}~ & \end{tabular}} &  
         \multicolumn{3}{c}{\begin{tabular}{ccc} & ~{\thirdc16.3}~ & \end{tabular}} &  
         \multicolumn{3}{c}{\begin{tabular}{ccc} &~{\secondc 17.4}~ & \end{tabular}} \\

         & & & & & & & & & & & & & & & & & & & & & & & & & & & & & \\[-10pt]
         & Palace & 509 & 
         \multicolumn{3}{c|}{\begin{tabular}{ccc} & ~{14.2}~ & \end{tabular}} &  
         \multicolumn{3}{c}{\begin{tabular}{ccc} & ~{\secondc11.7}~ & \end{tabular}} &  
         \multicolumn{3}{c}{\begin{tabular}{ccc} & ~{\thirdc10.8}~ & \end{tabular}} &  
         \multicolumn{3}{c}{\begin{tabular}{ccc} & ~{10.7}~ & \end{tabular}} & 
         \multicolumn{3}{c||}{\begin{tabular}{ccc} &~{\firstc14.3}~ & \end{tabular}} & 
         21.9k & 
        \multicolumn{3}{c}{\begin{tabular}{ccc} & ~{15.3}~ & \end{tabular}} & 
        \multicolumn{3}{c|}{\begin{tabular}{ccc} &~{\secondc12.3}~ & \end{tabular}} & 
         \multicolumn{3}{c}{\begin{tabular}{ccc} &~-~& \end{tabular}} &  
         \multicolumn{3}{c}{\begin{tabular}{ccc} &~{18.6}~& \end{tabular}} &  
         \multicolumn{3}{c}{\begin{tabular}{ccc} &~{\thirdc11.0}~& \end{tabular}} &  
         \multicolumn{3}{c}{\begin{tabular}{ccc} &~{\firstc13.3}~ & \end{tabular}} \\

         & & & & & & & & & & & & & & & & & & & & & & & & & & & & & \\[-10pt]
         & Temple & 302 & 
         \multicolumn{3}{c|}{\begin{tabular}{ccc} & ~{18.1}~ & \end{tabular}} & 
         \multicolumn{3}{c}{\begin{tabular}{ccc} & ~{\secondc15.7}~ & \end{tabular}} &  
         \multicolumn{3}{c}{\begin{tabular}{ccc} & ~{\thirdc11.8}~ & \end{tabular}} &  
         \multicolumn{3}{c}{\begin{tabular}{ccc} & ~{9.7}~ & \end{tabular}} &
         \multicolumn{3}{c||}{\begin{tabular}{ccc} &~{\firstc17.8}~ & \end{tabular}} & 
         17.5k & 
        \multicolumn{3}{c}{\begin{tabular}{ccc} & ~{19.6}~ & \end{tabular}} & 
        \multicolumn{3}{c|}{\begin{tabular}{ccc} & ~{\secondc 16.1}~ & \end{tabular}}  & 
         \multicolumn{3}{c}{\begin{tabular}{ccc} & ~-~ & \end{tabular}} &  
         \multicolumn{3}{c}{\begin{tabular}{ccc} & ~{11.9}~ & \end{tabular}} &  
         \multicolumn{3}{c}{\begin{tabular}{ccc} & ~{\thirdc 14.8}~ & \end{tabular}} &
         \multicolumn{3}{c}{\begin{tabular}{ccc} &~{\firstc18.3}~ & \end{tabular}}\\

         & & & & & & & & & & & & & & & & & & & & & & & & & & & & & \\[-10pt]
        \hhline{~-----------------------------------}
         & & & & & & & & & & & & & & & & & & & & & & & & & & & & & \\[-10pt]
         & Average & 348 & 
        \multicolumn{3}{c|}{\begin{tabular}{ccc} & ~{17.3}~ & \end{tabular}} &  
         \multicolumn{3}{c}{\begin{tabular}{ccc} & ~{\secondc15.0}~ & \end{tabular}} &  
         \multicolumn{3}{c}{\begin{tabular}{ccc} & ~{12.9}~ & \end{tabular}} &  
         \multicolumn{3}{c}{\begin{tabular}{ccc} & ~{\thirdc14.8}~ & \end{tabular}} & 
         \multicolumn{3}{c||}{\begin{tabular}{ccc} &~{\firstc16.9}~ & \end{tabular}} & 
         15.6k & 
        \multicolumn{3}{c}{\begin{tabular}{ccc} & ~{18.6}~ & \end{tabular}} &  
        \multicolumn{3}{c|}{\begin{tabular}{ccc} & ~{\secondc16.3}~ & \end{tabular}} & 
         \multicolumn{3}{c}{\begin{tabular}{ccc} & ~-~ & \end{tabular}} &  
         \multicolumn{3}{c}{\begin{tabular}{ccc} & ~{11.5}~ & \end{tabular}} &  
         \multicolumn{3}{c}{\begin{tabular}{ccc} & ~{\thirdc16.2}~ & \end{tabular}} &  
         \multicolumn{3}{c}{\begin{tabular}{ccc} &~{\firstc17.4}~ & \end{tabular}}\\

         & & & & & & & & & & & & & & & & & & & & & & & & & & & & & \\[-10pt]
        \hline
         & Time &      & 
         \multicolumn{3}{c|}{\begin{tabular}{ccc} & ~1h~ & \end{tabular}} &  
         \multicolumn{3}{c}{\begin{tabular}{ccc} & ~2min~ & \end{tabular}} &  
         \multicolumn{3}{c}{\begin{tabular}{ccc} & ~4min~ & \end{tabular}} & 
         \multicolumn{3}{c}{\begin{tabular}{ccc} & ~1h~ & \end{tabular}} & 
         \multicolumn{3}{c||}{\begin{tabular}{ccc} & ~3.5min~ & \end{tabular}}  &    & 
        \multicolumn{3}{c}{\begin{tabular}{ccc} & ~10h~ & \end{tabular}} &  
         \multicolumn{3}{c|}{\begin{tabular}{ccc} & ~2.1h~ & \end{tabular}} & 
         \multicolumn{3}{c}{\begin{tabular}{ccc} & ~    ~ & \end{tabular}} &  
         \multicolumn{3}{c}{\begin{tabular}{ccc} & ~27min~ & \end{tabular}} &  
         \multicolumn{3}{c}{\begin{tabular}{ccc} & ~2.8h~ & \end{tabular}} &  
         \multicolumn{3}{c}{\begin{tabular}{ccc} & ~59min~ & \end{tabular}}  \\
        \bottomrule
         
    \end{tabular}
    }
    \caption{\textbf{Tanks \& Temples.}  We show the pose accuracy via view synthesis with Nerfacto~\cite{nerfstudio} as PSNR in dB, and the reconstruction time. We color code in: \TCBTableBest {Best}, \TCBTableSecond{Second}, and \TCBTableThird{Third}. $^{\dagger}$Method needs sequential inputs.}%
        \vspace{-3mm}
    \label{tab:supp_t2_pose}%
\end{table*}

%% file: tables/table_7scenes_psnr_multi_recon.tex
\begin{table*}[!htp]
	\begin{center}
		\adjustbox{valign=t,width=\linewidth}{
			\sisetup{detect-all=true,detect-weight=true}
			\begin{tabular}{cc||ccc||cccc||ccccc}
				\toprule
				\multicolumn{2}{c||}{~} &%
				\multicolumn{3}{c||}{~Pseudo GT~} &%
				\multicolumn{4}{c||}{~All Frames~} &%
				\multicolumn{4}{c}{~200 Frames~} \\
				\hhline{~~-----------}
				~ &
				\hspace{-1pt}\parbox[t]{0mm}{\multirow{2}{*}{\rotatebox[origin=c]{90}{Frames\hspace{-12pt}}}} &%
				~Kinect~ &%
				~COLMAP~ &%
				\multicolumn{1}{c||}{~COLMAP~} &%
				\multicolumn{1}{c}{~DROID-~} &%
				\multicolumn{1}{c}{~ACE0~} &%
				\multicolumn{1}{c}{~ACE0~} &%
				\multicolumn{1}{c||}{~Ours} &%
				\multicolumn{1}{c}{~BARF} &%
				\multicolumn{1}{c}{~NoPE-} &%
				\multicolumn{1}{c}{~ACE0}&
				\multicolumn{1}{c}{~Ours}\\%
				&%

				~ &%
				Fusion &%
				(default)   &%
				\multicolumn{1}{c||}{(fast)} &%
				\multicolumn{1}{c}{~SLAM$^\dagger$} &
				\multicolumn{1}{c}{(default)} &%
				\multicolumn{1}{c}{(KF Init.)} &
				\multicolumn{1}{c||}{(50F)}&
				\multicolumn{1}{c}{\cite{lin2021barf}} &
				\multicolumn{1}{c}{NeRF}&
				\multicolumn{1}{c}{(default)}&
				\multicolumn{1}{c}{(50F)}\\

				\midrule 
				Chess      & 6k  & 19.6 & 23.6 & 23.5      
                & 19.3   & \secondc23.3 & \thirdc23.0 & \firstc23.4 &       
                \thirdc 12.8 & 12.6 & \firstc22.7 & \secondc21.8\\
				Fire       & 4k  & 19.2 & 22.6 & 22.6      
                & \thirdc13.0   & \secondc22.3 & \secondc22.3 & \firstc22.8 &    
                \thirdc12.7 & 11.8 & \secondc22.1 &  \firstc24.4\\
				Heads      & 2k  & 17.0 & 18.8 & 18.9       
                & 17.6   & \secondc18.8 & \firstc19.1 &\thirdc18.5 &     
                10.7 & 11.8 & \secondc19.9 & \firstc20.2\\
				Office     & 10k & 18.9 & 21.4 & 21.6        
                & failed & \secondc21.1 & \firstc21.5 &\thirdc 20.9 &    
                \thirdc11.9 & 10.9 & \firstc19.8 & \secondc19.4\\
				Pumkin     & 6k  & 19.9 & 24.1 & 23.8     
                & 18.3   & \secondc24.1 & \thirdc23.8 & \firstc24.5 &    
                \thirdc 19.6 & 14.2 & \secondc24.7 & \firstc25.0\\
				RedKitchen & 12k & 17.6 & 21.4 & 21.4      
                & 10.9   & \secondc 20.8 & \firstc20.9 & \thirdc 19.9 &   
                \thirdc11.6 & 11.2 & \secondc18.9 & \firstc20.0\\
				Stairs     & 3k  & 19.0 & 16.7 & 21.0      
                & 13     & \thirdc17.7 & \secondc 19.9 & \firstc20.6 &    
                15.8 & \thirdc15.9 & \secondc18.8 & \firstc20.8\\
                \hline
				Average    & 6.5k& 18.7 & 21.2 & 21.8       
                & N/A    &\thirdc  21.2 & \firstc21.5 & \firstc21.5 &     
               \thirdc 13.6 & 12.6 & \secondc21.0 & \firstc21.8\\
				Avg. Time  & &realtime &  38h  & 13h     
                & 18min  & 1h   & 7min & 25min &    
                8.5h & 47h  & 27min& 3min \\
				\bottomrule
			\end{tabular}
		}
		\caption{\textbf{7-Scenes.} We show the pose accuracy via view synthesis with Nerfacto~\cite{nerfstudio} as PSNR in dB, and the reconstruction time.  We color code in: \TCBTableBest {Best}, \TCBTableSecond{Second}, and \TCBTableThird{Third}. Our method takes 50 frames as anchor images only, achieves SOTA performance. For some competitors, we had to sub-sample the images due to their computational complexity (right side). $^{\dagger}$Method needs sequential inputs. 
		}
		\label{tab:7scenes_psnr}\end{center}
		\end{table*}